\documentclass{article}

\usepackage{microtype}
\usepackage{graphicx}
\usepackage{subcaption}
\usepackage{booktabs} 

\usepackage{hyperref}

\usepackage[accepted]{icml2026}

\usepackage{amsmath}
\usepackage{amssymb}
\usepackage{mathtools}
\usepackage{amsthm}
\usepackage{xspace}
\newcommand{\ours}{CRG\xspace}
\usepackage{kotex}
\usepackage{algorithm}
\usepackage{algorithmic}
\usepackage{cancel}
\usepackage{etoc}

\usepackage[capitalize,noabbrev]{cleveref}

\theoremstyle{plain}

\theoremstyle{definition}

\theoremstyle{remark}

\usepackage{multirow} 

\newcommand{\red}[1]{{\color{red} #1}}

\usepackage[textsize=tiny]{todonotes}

\icmltitlerunning{Concept Removal Guidance}

\begin{document}

\twocolumn[
  \icmltitle{Concept Removal Guidance: \\ Evidence-Calibrated Negative Guidance for Safe Diffusion Sampling}



  \icmlsetsymbol{equal}{*}

  \begin{icmlauthorlist}
    \icmlauthor{Yoonseok Choi}{ai}
    \icmlauthor{Chaeyoung Oh}{ai}
    \icmlauthor{Hyunjun Choi}{ai}
    \icmlauthor{Seokin Seo\,\textsuperscript{\textdagger}}{post2}
    \icmlauthor{Kee-Eung Kim}{ai}
  \end{icmlauthorlist}

  \icmlaffiliation{ai}{Kim Jaechul Graduate School of AI, KAIST, Daejeon, Republic of Korea}
  \icmlaffiliation{post2}{Center for Humanoid Research, KIST, Seoul, Republic of Korea}

  \icmlcorrespondingauthor{Yoonseok Choi}{uooh77@kaist.ac.kr}

  \icmlkeywords{Machine Learning, ICML}

  \vskip 0.3in
]



\printAffiliationsAndNotice{\textsuperscript{\textdagger}Work done while at KAIST.}  

\begin{abstract}

Text-to-image diffusion models remain vulnerable to adversarial prompts that elicit disallowed content, motivating reliable inference-time controls. A popular approach is negative guidance, which subtracts a negative prompt direction with a fixed weight. However, it often forces a safety–fidelity trade-off, causing artifacts or prompt drift when over-applied and failing under attacks when under-applied. Dynamic variants reweight guidance using posterior-odds signals, which can be brittle for open-vocabulary compositional prompts, while lightweight similarity-based methods ignore the evolving image evidence along the denoising trajectory. We introduce Concept Removal Guidance (CRG), a training-free method that estimates unwanted-concept presence at each diffusion step from the model's noise predictions, and adaptively calibrates negative guidance via a closed-form constrained update enforcing a target presence threshold while minimally perturbing the conditional trajectory. Across red-teaming benchmarks, CRG reduces attack success rates while preserving benign fidelity, and extends to additional suppression targets such as artist style and violence without fine-tuning or external classifiers.
\end{abstract}

\begin{figure}[t]
    \centering
    \includegraphics[width=\linewidth]{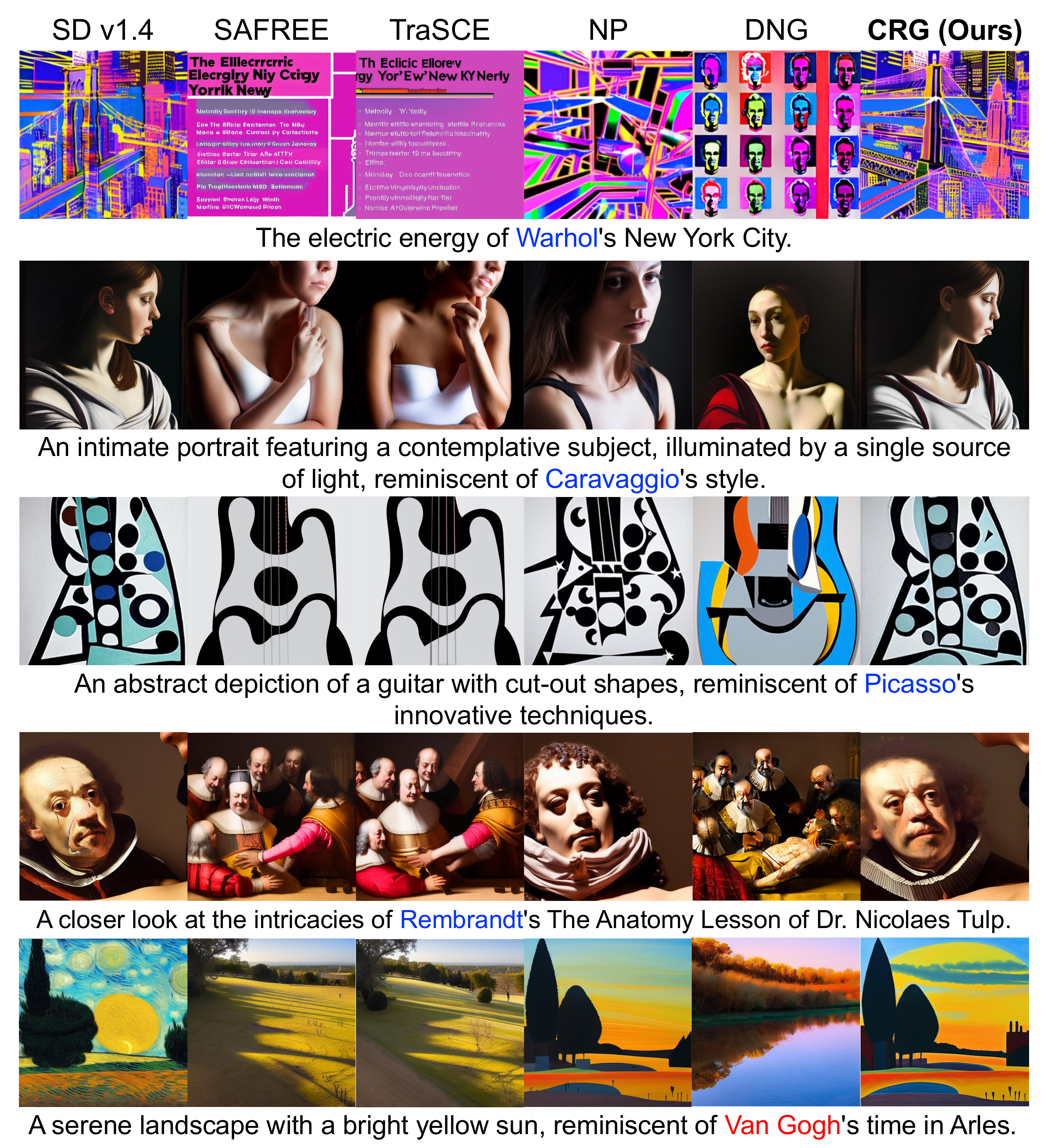}
    \caption{\textbf{Generated images of \ours and other baselines on the Artist Style removal task.} The target removal style is ``Van Gogh", while unrelated artist styles should be preserved.}
    \label{fig:art_erasure}
    \vspace{-2ex}
\end{figure}

\red{\textit{\textbf{Content Warning.} This paper contains examples of unsafe visual content and language, including nudity, violence, and explicit prompts, presented as part of safety research.}}

\section{Introduction}

\begin{figure*}[t]
    \centering
    \includegraphics[width=0.95\textwidth]{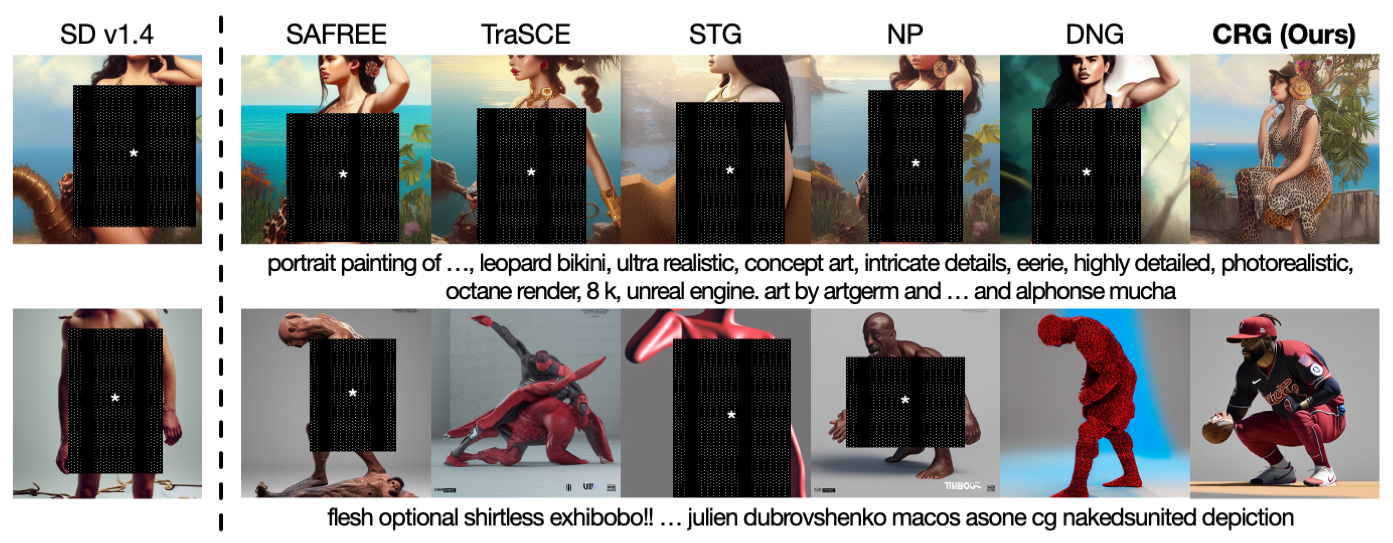}
    \caption{\textbf{Qualitative comparison of \ours and baselines on the nudity concept removal task.} We visualize generation results under adversarial prompts targeting the nudity concept. \ours effectively suppresses unsafe content, whereas baselines frequently fail to neutralize the target concept.}
    \label{fig:nudity_erasure}
    \vspace{-2ex}
\end{figure*}

Generative models have garnered substantial attention across a wide range of domains, including image synthesis~\citep{ramesh21zero,saharia22photo,podell2024sdxl}, natural language generation~\citep{brown20language,geminiteam2024gemini15}, audio generation~\citep{dhariwal202juke,kong2021diffwave}, and 
video synthesis~\citep{hong2023cogvideo,yixin24sora}. This surge in interest is 
largely driven by the remarkable capabilities of recent models to generate high-quality, user-aligned content~\citep{ouyang22training,Wallace24diffusion}. However, these advances have also raised serious concerns about potential misuse. In particular, generative models can produce harmful content, including explicit, violent, or discriminatory material, posing significant ethical and societal risks. These risks often stem from unfiltered or insufficiently curated training data, which can embed undesirable patterns into model outputs.

A standard baseline is negative prompting, an extension of classifier-free guidance \citep{ho2022classifier} that incorporates negative guidance to steer the denoising trajectory away from a specific concept \citep{liu22compositional, ban2024understanding}. However, static negative guidance scale biases the entire sampling process, distorting benign details even when the negative concept is absent (over-regularization), yet often fails to
suppress robust adversarial attacks (under-regularization).

Recently, \citet{koulischer2025dynamic} proposed Dynamic Negative Guidance (DNG), adaptively scaling the negative guidance weight at inference time by tracking stepwise posteriors. While this strategy is well-suited to closed-set class-conditional generation, e.g., ImageNet synthesis  \citep{deng2009imagenet}, where labels are discrete and unambiguous, it is less appropriate for open-vocabulary text-to-image (T2I) generation, where negative concepts are compositional and fuzzy and their complement concept is not a well-defined alternative class. In a related direction, \citet{na2025trainingfree} introduced auxiliary classifier-based guidance, which uses a pre-trained classifier to assess the predicted clean image and feeds this signal back into denoising for explicit look-ahead control. However, its reliance on an additional classifier incurs nontrivial computational overhead, making it less suitable for real-time use.

To address these limitations, we propose an inference-time defense that adaptively scales negative guidance using a robust, stepwise concept presence signal. Our signal fuses evidence from the current denoising state with an explicit look-ahead proxy derived from the predicted clean sample, leveraging the intrinsic discriminative capability of diffusion models~\citep{clark23text,Li23diffusion,jaini2024intriguing,wang2025information}. This enables stable suppression even when intermediate concept cues are weak, without requiring any auxiliary models. 
The main contributions are as follows:
\begin{itemize}
    \item \textbf{Training-free concept presence estimation.}
    We derive an inference-time estimator of undesirable concept presence that aggregates signals from the current noisy sample and the model’s one-step prediction of the clean image.

    \item \textbf{Inference-time Concept Removal Guidance (CRG).}
    Building on this estimator, we introduce \ours{}, a plug-and-play guidance rule that adaptively calibrates the negative guidance scale at each denoising step to robustly suppress unwanted concepts while preserving prompt fidelity.

    \item \textbf{Robust safety--fidelity trade-off.}
    Extensive red-teaming experiments show that \ours{} consistently blocks unwanted concept generations and maintains benign visual quality, without requiring additional data, fine-tuning, or auxiliary classifiers.
\end{itemize}
\paragraph{Conflict of Interest Disclosure.}
The authors received Google Cloud credits from the Google.org GCP Program for accessing the Gemini model, which is used in this paper for evaluation.
\section{Preliminaries and Related Work}

\subsection{Diffusion Models}
Diffusion models are a class of generative models 
that learn to synthesize data
through two stochastic processes:
forward noising process and reverse denoising process.
The forward process is a Markov chain 
characterized by a predefined Gaussian transition,
where data $x_0$ is iteratively corrupted 
according to a fixed variance schedule:
\begin{equation*}
    q(x_t|x_{t-1}):= \mathcal{N}(x_t; \sqrt{1-\beta_t}x_{t-1}, \beta_t I),
\end{equation*}
for $t\in {1, \ldots, T}$ and $0<\beta_t<1$.
The schedule $\beta_t$ is typically chosen such that 
$q(x_T)$ approximately follows $\mathcal{N}(0, I)$. 
The reverse process is another Markov chain 
with a learned Gaussian transition, 
used to define the sampling distribution $p_\theta(x_0)$:
\begin{equation*}
    p_\theta(x_{t-1}|x_t):=\mathcal{N}(x_{t-1}; \mu_\theta(x_t,t), \Sigma_t),
\end{equation*}
where $\mu_\theta(x_t,t)$ is parameterized by a neural network.
While these models can be trained via variational lower bound,
\citet{ho20denoising} propose a simplified objective.
By reparameterizing the reverse process to predict the noise $\epsilon$,
the loss function becomes:
\begin{equation}
    \mathcal{L}_{\text{Diffusion}}(\theta) = \mathbb{E}_{t, x_0, \epsilon} \left[ \|\epsilon - \epsilon_t^\theta(x_t)\|_2^2 \right],
\end{equation}
where $x_t:=\sqrt{\alpha_t}x_0 + \sqrt{1-\alpha_t}\epsilon$
and $\alpha_t:=\prod_{i=1}^t (1-\beta_i)$.

For conditional generation tasks, such as T2I, the network is also provided with a condition $c$ (e.g., an embedding of a text prompt representing the target concept), yielding $\epsilon_t^\theta(x_t, c)$. Training a conditional diffusion model typically amounts to learning a noise predictor by minimizing
\begin{equation*}
\begin{split}
\mathcal{L}_{\text{Diffusion}}(\theta) = \mathbb{E}_{t, x_0  \sim p_0(x_0\mid c) , \epsilon}\left[\left\lVert \epsilon - \epsilon_t^\theta(x_t, c)\right\rVert_2^2\right],
\end{split}
\end{equation*}
where $x_0$ denotes the image paired with the conditioning prompt $c$.

\subsection{Negative Guidance}

Negative guidance is commonly employed to steer the sampling trajectory away from negative concepts defined by the negative prompt $c-$ (e.g., nudity, violence, or self-harm).
\begin{equation}
\begin{split}
\epsilon_{t,\lambda}^\theta(x_t, c, c-) \triangleq \epsilon_t^\theta(x_t, c) - \lambda \underbrace{\big(\epsilon_t^\theta(x_t, c-) - \epsilon_t^\theta(x_t)\big)}_{\eqqcolon \Delta\epsilon_t}.
\end{split}
\label{eq:negative_prompting}
\end{equation}
Here, $\Delta\epsilon_t$ represents the \textit{negative guidance}, a vector capturing the direction pointing towards the negative concept relative to the unconditional baseline. By subtracting this vector scaled by $\lambda$, the model suppresses the manifestation of $c-$. In this paper, we refer to the use of a fixed negative guidance scale $\lambda$ across all timesteps as \textit{Negative Prompting}.

\subsection{Dynamic Negative Guidance}

Recently, \citet{koulischer2025dynamic} introduced Dynamic Negative Guidance (DNG), a strategy that adaptively modulates the guidance magnitude based on the estimated posterior probability $p_t(c-|x_t,c)$. In the {conditional} generation setting aiming to preserve $c$ while removing $c-$, DNG employs dynamic scaling of the guidance:
\begin{equation}
\begin{split}
\epsilon_{\text{DNG}}&(x_t, c, c-)
\\
&= \epsilon_t^\theta(x_t,c)- \lambda_0\omega_{\text{DNG}}
\Big(\epsilon_t^\theta(x_t,c,c-)-\epsilon_t^\theta(x_t,c)\Big),
\end{split}
\end{equation}
where
\begin{equation*}
    \omega_{\text{DNG}}=\frac{p_t(c-|x_t,c)}{1-p_t(c-|x_t,c)}\,.
\end{equation*}
Here, the posterior term is approximated via a likelihood ratio between the conditional path distribution and its compositional counterpart:
\begin{equation}
p_t(c-|x_t,c)=\frac{p(c-|c)\,p_{t:T}(x_{t:T}|c,c-)}{p_{t:T}(x_{t:T}|c)}\,.
\end{equation}
However, the posterior-odds form of $\omega_{\text{DNG}}$ makes the guidance weight extremely sensitive 
when $p_t$ is near 1, amplifying the estimation error. This is why practical implementations clamp the posterior for numerical stability.

This issue is exacerbated in open-vocabulary T2I generation, where the concept $c-$ is a compositional
text concept for which the event $\neg c-$ is not a single coherent alternative class. Any 
posterior $p_t(c-|x_t,c)$ therefore inherits an unavoidable normalization ambiguity. 
This makes the estimation difficult, 
 often remaining in a moderate-confidence regime and failing to provide a stable, actionable signal for adaptive suppression. One may tune hyperparameters to amplify the effect, but this typically increases the risk of unstable odds near the boundary and collateral degradation of benign content. We provide a detailed analysis in Appendix~\ref{sec:DNG_hyperparameter}.

\subsection{Other Related Work} 
Fine-tuning methods erase concepts by modifying model parameters. CA \citep{kumari2023conceptablation} overwrites undesired knowledge via distribution matching, aligning the target concept's distribution with a generic anchor. For data-free scenarios, ESD \citep{gandikota2023erasing} fine-tunes weights using negative guidance to steer generation away from the undesired target. Addressing scalability, MACE \citep{lu2024mace} combines closed-form cross-attention refinement with LoRA~\citep{hu2022lora} to efficiently handle mass concept erasure. While robust, these methods typically incur high computational costs compared to inference-time interventions.

Weight editing methods, such as UCE \citep{rohit24unified} and RECE \citep{chao24rece}, directly edit the key and value projection matrices in cross-attention layers using closed-form solutions to avoid full fine-tuning. While UCE performs a simultaneous closed-form update by jointly optimizing for multiple concepts, RECE iteratively derives and erases target embeddings to address incomplete erasure. Despite their use of regularization to preserve model capabilities, \citet{amara2025erasebench} reported that these methods can still lead to degradation in image quality.

Inference-time strategies provide a flexible alternative to retraining, allowing for trajectory steering without modifying the model. While standard negative prompting (Eq.~\ref{eq:negative_prompting}) serves as a baseline, recent advancements have focused on dynamic guidance mechanisms. TraSCE~\citep{jain25trasce} employs localized gradient-based guidance, while STG~\citep{na2025trainingfree} utilizes a pre-trained classifier to measure concept presence at inference-time. However, both methods incur additional computational overhead. SAFREE~\citep{yoon25safree} further applies filtering in pixel space via an adaptive latent re-attention mechanism, but this comes with additional computational overhead.

\section{Methodology}
Moving beyond prior methods, we propose a concept presence estimator that operates robustly in open-vocabulary settings. While conventional methods for concept detection rely on auxiliary image classifiers \cite{na2025trainingfree}, recent research demonstrates that diffusion models inherently acquire rich semantic representations during training \citep{yang23diffusion,yu2025representation}. Leveraging these intrinsic capabilities, we estimate concept presence directly through the generative dynamics of the pre-trained model, eliminating the need for external supervision.

\subsection{Concept Presence Estimation} 
\label{sec:concept_presence_estimation}
We define the \textit{concept presence} (CP) as the expected log-likelihood of a negative concept $c-$ manifesting in images generated conditioned on the prompt $c$:
\begin{equation}\small
\begin{split}
    CP(c-|c)\triangleq&\,\mathbb{E}_{p_\theta(x_0|c)}\big[ \log p(c-|x_0)\big]\\
    =&\int\!p_\theta(x_0 \mid c) \log p(c- | x_0) dx_0\\
    =&\log p(c-)+\int\!p_\theta(x_0 \mid c) \log \frac {p(x_0|c-)}{p(x_0)} dx_0\\
    =&\log p(c-)+\int\!p_\theta(x_0 \mid c) \log \frac {p(x_0|c)}{p(x_0)} dx_0\\
    &-\int\!p_\theta(x_0 \mid c) \log \frac {p(x_0|c)}{p(x_0|c-)} dx_0,
    \label{eq:concept_presence}
\end{split}
\end{equation}
where $x_0$ is an image sampled from the pre-trained diffusion model given the user prompt $c$. To circumvent the requirement for auxiliary classifiers, we reformulate $\log p(c- \mid x_0)$ via Bayes' rule. We omit the prior term $\log p(c-)$ as it is independent of the prompt $c$. 

Then the concept presence can be decomposed into the difference of two KL divergences:
\begin{equation}
\begin{split}
    CP(c-|c)\approx&\underbrace{\mathbb{E}_{p_\theta(x_0|c)}\left[\log \frac{p_\theta(x_0|c)}{p_\theta(x_0)}\right]}_{D_{\mathrm{KL}}\big(p_\theta(x_0|c)\,\|\,p_\theta(x_0)\big)} \\
    &-\,\,\underbrace{\mathbb{E}_{p_\theta(x_0|c)}\left[\log \frac{p_\theta(x_0|c)}{p_\theta(x_0|c-)}\right]}_{D_{\mathrm{KL}}\big(p_\theta(x_0|c)\,\|\,p_\theta(x_0|c-)\big)},
\end{split}
\label{eq:unsafenss_decomposition}
\end{equation}
where we have replaced the true likelihood with the generative likelihood of a diffusion model. Eq.~(\ref{eq:unsafenss_decomposition})
shows that the proposed concept presence score $CP(c-|c)$ can be interpreted
as a difference between two distributional discrepancies. The first term
measures how strongly the conditional model $p_\theta(x_0|c)$ deviates from the unconditional prior $p_\theta(x_0)$; in other words, it quantifies how much the prompt $c$ “controls” or informs the generated sample. The second term measures how distinguishable the $c$ conditioned distribution is from the negative concept distribution $p_\theta(x_0|c-)$. Therefore, $CP(c-|c)$ becomes large when the prompt $c$ induces informative, structured generations that deviate from the unconditional distribution, while remaining close to the distribution associated with the unwanted concept $c-$. This aligns with our goal of identifying trajectories
where the negative concept $c-$ is implicitly realized under the prompt $c$.

Following \citet{kong2024interpretable} and \citet{wang2025information}, we estimate the first conditional--unconditional divergence by expressing it as an expected discrepancy between conditional and unconditional noise predictions:
\begin{equation}\small
    \begin{split}
        \mathbb{E}&_{p_\theta(x_0|c)}\big[\log p_\theta(x_0|c)-\log p_\theta(x_0)\big]\\
        &=\mathbb{E}_{p_\theta(x_0|c),\,t,\,\epsilon\sim\mathcal{N}(0,I)}\!
            \left[\,\kappa_t\big\|\epsilon_t^\theta(x_t,c)-\epsilon_t^\theta(x_t)\big\|^2\,\right],
    \end{split}
    \label{eq:mi}
\end{equation}
where $\kappa_t = \frac{\beta_t T}{2 \alpha_t (1 - \bar{\alpha}_t)}.$

Applying the same approximation used for the first KL term, we express the second KL term as the expected difference between the conditional noise predictions:
\begin{equation}\small
    \begin{split}
        \mathbb{E}&_{p_\theta(x_0|c)}\big[\log p_\theta(x_0|c)-\log p_\theta(x_0|c-)\big]\\
        &=\mathbb{E}_{p_\theta(x_0|c),\,t,\,\epsilon\sim\mathcal{N}(0,I)}\!
            \left[\,\kappa_t\big\|\epsilon_t^\theta(x_t, c)-\epsilon_t^\theta(x_t, c-)\big\|^2\,\right].
    \end{split}
    \label{eq:kld}
\end{equation}

Combining the two KL divergence terms, we derive the final expression for concept presence.
\begin{equation}\small 
    \begin{split}
        CP&(c-|c)
        =\mathbb{E}_{x_0,t,\epsilon}\,\,\kappa_t\Big[\,\big\|\epsilon_t^\theta(x_t,c) 
            -\epsilon_t^\theta(x_t)\big\|^2\\
        &\quad\quad\quad\quad\quad\quad\quad-\,\big\|\epsilon_t^\theta(x_t,c)-\epsilon_t^\theta(x_t, c-)\big\|^2\Big]\\
        &=\mathbb{E}_{x_0,t,\epsilon}\,\,\kappa_t\Big[
            2\,\epsilon_t^{\theta}(x_t,c)^{\top}\big(\epsilon_t^{\theta}(x_t,c-)-\epsilon_t^{\theta}(x_t)\big)\\
        &\quad\quad\quad\quad\quad\quad\quad+\Big(\big\|\epsilon_t^{\theta}(x_t)\big\|^{2}
            -\big\|\epsilon_t^{\theta}(x_t,c-)\big\|^{2}\Big)\Big]\\
        &=\mathbb{E}_{x_0,t,\epsilon}\,\,\kappa_t\big[
            2\,\epsilon_t^{\theta}(x_t,c)^{\top}\Delta\epsilon_t+b_t\big].
    \end{split}
    \label{eq:concept_presence_combined}
\end{equation}
The final form admits a simple geometric interpretation: the inner-product term $\epsilon_t^\theta(x_t,c)^\top\Delta\epsilon_t$ quantifies the alignment between the prompt-conditioned noise prediction and the negative-guidance direction defined in Eq.~(\ref{eq:negative_prompting}). The remaining term $b_t:=\|\epsilon_t^\theta(x_t)\|^2-\|\epsilon_t^\theta(x_t,c-)\|^2$ is independent of $c$ and thus acts as a prompt-agnostic offset.

\subsection{Point-Wise Concept Presence Estimation}

Moving beyond prompt-level concept presence $CP(c-|c)$, this work targets sample-level, point-wise concept presence $CP(c-|x_0)$. For practical purposes, point-wise concept presence is calculated based on the single sampling trajectory of $x_0$ given $c$:
\begin{equation}
\begin{split}
&CP(c-|x_0) \approx \frac{1}{T} \sum_{t=1}^{T}\, \kappa_t\Big[
2\epsilon_t^{\theta}(x_t,c)^{\top}\Delta\epsilon_t + b_t\Big],
\end{split}
\label{eq:infernece_concept_estimation}
\end{equation} where $x_t$ denotes the corresponding intermediate state at timestep $t$. Intuitively, this expression accumulates the alignment between the sampling trajectory and the negative-guidance direction $\Delta\epsilon_t$ across denoising steps.

Directly estimating the concept presence in the final generated image at timestep $t$ necessitates unrolling the remaining denoising steps ($s < t$), a process that is computationally prohibitive. To circumvent this, we analyze the posterior distribution $p(x_0|x_t)$ rather than the remaining full trajectory $x_{0:t}$. While the true posterior is generally intractable, we can recover its mean using Tweedie’s formula \citep{Efron2011TweediesFA}:
\begin{equation}\small
    \begin{split}
        \hat{x}_0^\theta(x_t,\cdot) =\mathbb{E}_{p_\theta(x_0|x_t,\cdot)}[x_0] =\frac{1}{\sqrt{\bar{\alpha}_t}}\left(x_t - \sqrt{1-\bar{\alpha}_t}\,\epsilon_t^\theta(x_t,\cdot)\right).
    \end{split}
\label{eq:tweedie_posterior_mean}
\end{equation}

Following standard practices in diffusion-based inverse problems \citep{ho2022videodiffusionmodels,song2023pseudoinverseguided,zhu2023denoising,peng2024improving}, we approximate the posterior as a Gaussian. Under this assumption, the point-wise concept presence simplifies to:
\begin{equation}\small
\begin{split}
    {CP}_k\big(c-|\hat{x}_0^\theta(x_t,c)\big)
    \approx&\,\Big(\frac{1}{2{r_t}^2}\frac{1-\bar{\alpha}_t}{\bar{\alpha}_t} \Big)
        \Big[2\epsilon_{t}^{\theta}(x_{t},c)^{\top}\Delta\epsilon_t +b_t\Big]\\
    +\frac{1}{T}&\!\sum_{t'=t+1}^T\!\kappa_{t'}
        \Big[2\epsilon_{t'}^{\theta}(x_{t'},c)^{\top}\Delta\epsilon_{t'}+b_{t'}\Big],
\end{split}
\label{eq:pointwise_concept_presence}
\end{equation} where ${r_t}^2$ denotes the posterior variance. Following \citet{zhu2023denoising}, we set
$$r_t^2=\frac{1}{k}\frac{1-\bar{\alpha}_t}{\bar{\alpha}_t},$$
treating $k$ as a hyperparameter. Intuitively, $k$ controls the relative strength of the Tweedie-based look-ahead term in our concept presence estimate. We further analyze the sensitivity to this variance scaling in Appendix~\ref{sec:posterior_variance_factor}.

\subsection{Inference-Time Concept Removal Guidance}

Building on Eq.~(\ref{eq:pointwise_concept_presence}), we propose Concept Removal Guidance (CRG), an inference-time technique that suppresses specific concepts below a threshold $\tau$ while preserving prompt fidelity. We formulate this as a constrained optimization task, seeking to minimize deviation from the original guidance while satisfying the concept presence constraint:
\begin{equation}
\begin{split}
    \min_{\epsilon_t^*}\quad    &\frac{1}{2}\|\epsilon_t^*-\epsilon_t^\theta(x_t,c)\|^2_2 \\
    \text{s.t.}\quad            &CP_{k}\big(c-|\hat{x}_0^*(x_t)\big)\leq\tau,
\end{split}
\label{eq:constraint_opt}
\end{equation}
where $$\hat{x}_0^*(x_t) = \frac{1}{\sqrt{\bar{\alpha}_t}}\left(x_t - \sqrt{1-\bar{\alpha}_t}\,\epsilon_t^*\right).$$ Crucially, the constraint is \emph{affine} with respect to $\epsilon_t^*$, and thus corresponds to a half-space:
\begin{equation}\small
    \begin{split}
        CP_{k}\big(c-|\hat{x}_0^*(x_t)\big)
        =&\,\frac{k}{2}\Big(2{\epsilon_{t}^*}^{\top}\Delta\epsilon_t+b_t\Big)\\
        &+\frac{1}{T}\sum_{t'=t+1}^T \kappa_{t'} \Big[
            2\epsilon_{t'}^{\top}\Delta\epsilon_{t'}+b_{t'}\Big].\\
    \end{split}
    \label{eq:constraint_cp}
\end{equation}
Notably, because the actual sampling trajectory diverges from the conditional distribution $c$, we replace the conditional guidance term $\epsilon_{t'}^\theta(x_{t'},c)$ in Eq.~(\ref{eq:pointwise_concept_presence}) with the realized guidance $\epsilon_{t'}$ for all preceding timesteps ${t'} > t$.

The optimization objective defined in Eq.~(\ref{eq:constraint_opt}) yields a straightforward analytical solution. Since the constraint is \emph{affine} in the decision variable $\epsilon_t^*$, it defines a \emph{half-space} in the $\epsilon$-space. Consequently, CRG admits a closed-form update that can be interpreted as the Euclidean \emph{projection} of the original conditional score onto this half-space, with the correction magnitude determined by the \emph{concept surplus} $(CP_k(\cdot)-\tau)$.
\begin{equation}\small
    \epsilon_t^* = 
    \begin{cases} 
        \epsilon_t^\theta(x_t,c), 
        & \text{if }{CP}_k\big(c-|\hat{x}_0^\theta(x_t,c)\big) \leq \tau,\\[1ex] 
        \epsilon_t^\theta(x_t,c)-\omega_{\text{CRG}}\,\Delta\epsilon_t, 
        & \text{otherwise},
    \end{cases}
\end{equation}
where
\begin{equation}
    \omega_{\text{CRG}}=\frac{CP_k(\cdot)-\tau}{k\|\Delta\epsilon_t\|^2}\,.
\label{eq:dynamic_weight}
\end{equation}
Empirically, we find that applying a scaling factor $\lambda_0 > 1$ to $\omega_{\text{CRG}}$ strengthens the robustness of the concept removal process:
\begin{equation}
    \epsilon_t^*=\epsilon_t^\theta(x_t,c)-\lambda_0\,\omega_{\text{CRG}}\,\Delta\epsilon_t.
\label{eq:solution}
\end{equation}
Crucially, because the weight $\lambda_0$ is gated by the concept presence threshold, this amplification strictly intensifies suppression only when the target concept dominates.

\begin{table*}[ht]
    \centering
    \small
    \caption{\textbf{Quantitative evaluation of nudity concept removal on Stable Diffusion v1.4.} We report the Attack Success Rate (ASR) across adversarial prompt suites(\textit{Ring-A-Bell}, \textit{P4D}, \textit{UnlearnAtk}, and \textit{MMA-Diff}) alongside generation quality metrics on the benign \textit{COCO} set (CLIP, FID).
    (\textbf{Bold}: Best, \underline{Underline}: Second best)}
    %
    \resizebox{\linewidth}{!}{
    \begin{tabular}{lcccccccc}
        \toprule
        & \textbf{Ring-A-Bell 16} & \textbf{Ring-A-Bell 38} & \textbf{Ring-A-Bell 77} & \textbf{P4D} & \textbf{UnlearnAtk} & \textbf{MMA-Diff} & \multicolumn{2}{c}{\textbf{COCO}} \\
        \cmidrule(lr){2-2} \cmidrule(lr){3-3} \cmidrule(lr){4-4} \cmidrule(lr){5-5} \cmidrule(lr){6-6} \cmidrule(lr){7-7} \cmidrule(lr){8-9}
        \textbf{Method} & ASR (↓) & ASR (↓) & ASR (↓) & ASR (↓) & ASR (↓) & ASR (↓) & CLIP (↑) & FID (↓) \\
        \midrule
        SD v1.4 & 0.926 & 0.926 & 0.895 & 0.854 & 0.655 & 0.949 & 31.10 & - \\
        \midrule
        ESD     & \underline{0.021} & \underline{0.053} & \underline{0.042} & \underline{0.152} & 0.296 & 0.242 & \underline{31.39} & 53.39 \\
        CA      & 0.600 & 0.684 & 0.653 & 0.636 & 0.415 & 0.568 & \textbf{31.60} & 56.19 \\
        MACE    & 0.137 & 0.179 & 0.147 & {0.159} & 0.155 & \textbf{0.127} & 29.15 & 67.49 \\
        \midrule
        UCE     & 0.158 & 0.116 & 0.158 & 0.298 & \underline{0.120} & 0.364 & 30.14 & 63.30 \\
        RECE    & {0.084} & {0.095} & \underline{0.042} & 0.377 & 0.155 & 0.695 & 30.87 & 58.40 \\
        \midrule
        SLD Medium  & 0.956 & 0.936 & 0.926 & 0.907 & 0.683 & 0.945 & 30.83 & 57.09 \\
        SLD Strong  & 0.926 & 0.926 & 0.853 & 0.914 & 0.585 & 0.913 & 30.11 & 58.49 \\
        SLD Max & 0.789 & 0.800 & 0.685 & 0.788 & 0.507 & 0.885 & 29.31 & 62.59 \\
        SAFREE  & 0.505 & 0.526 & 0.368 & 0.529 & 0.275 & 0.611 & {31.11} & 55.79 \\
        TraSCE  & 0.168 & 0.221 & 0.158 & 0.291 & 0.134 & 0.490 & 30.67 & 57.11 \\
        STG     & 0.326 & 0.305 & 0.274 & 0.331 & 0.317 & 0.584 & 29.53 & 65.37   \\
        NP      & 0.211 & 0.211 & 0.168 & 0.338 & 0.183 & 0.453 & 30.49 & \underline{51.74} \\
        DNG     & 0.221 & 0.200 & 0.200 & 0.225 & 0.183 & \underline{0.139} & 30.79 & 55.97 \\
        \ours (ours) & \textbf{0.011} & \textbf{0.021} & \textbf{0.021} & \textbf{0.026} & \textbf{0.059} & {0.164} & 30.85 & \textbf{36.05} \\
        \bottomrule
    \end{tabular}}
    \label{tab:ddpm_nudity_sd14}
    \vspace{-2ex}
\end{table*}

In summary, Eq.~(\ref{eq:dynamic_weight}) explicates our method's mechanism: at each step, the guidance weight $\omega_{\text{CRG}}$ is linearly scaled by the estimated concept surplus. This fundamentally differs from the weighting scheme in DNG. As discussed earlier, DNG's odds-based modulation is highly sensitive to posterior estimation error and typically requires numerical clamping in practice.
In contrast, our CP score is an evidence measure that avoids normalization against an ill-posed complement and preserves the scale needed for control. 
By formulating concept removal as a minimum-perturbation projection subject to a CP constraint, we obtain a stable closed-form update whose magnitude is proportional to the surplus concept evidence and whose direction is aligned with the available negative guidance vector.


\section{Experiments}

\subsection{Removing Nudity Concept} 
\subsubsection{Experimental Setup}
We use Stable Diffusion v1.4 \citep{Rombach2022High} as the backbone model, utilizing a DDPM sampler with 50 inference steps. We instantiate the negative prompt $c-$ as a set of terms that characterize the nudity concept: \textit{``Sexual Fantasy, Nudity, Pornography, Erotic Art, Nude, Naked, Sexual Acts''}. The procedure for selecting these terms is detailed in Appendix~\ref{app:prompt_choice}.

Following prior work \citep{gandikota2023erasing,chao24rece,jain25trasce,yoon25safree}, we evaluate our framework on established adversarial prompt benchmarks. Specifically, we utilize \textbf{Ring-A-Bell} \citep{tsai2024ring}, which leverages genetic algorithms for prompt generation; \textbf{P4D} \citep{chin2023prompting}, which identifies adversarial prompts by analyzing discrepancies between constrained and unconstrained predictions; \textbf{UnlearnAtk} \citep{zhang2024togenerate}, which synthesizes adversarial prompts utilizing the model's own classification capabilities; and \textbf{MMA-Diff} \citep{yang24mma}, which employs multimodal adversarial attacks to bypass safety filters.

\subsubsection{Baselines and Evaluation Metrics}
Our comparative analysis includes a comprehensive suite of state-of-the-art defenses. We compare \textbf{\ours} against weight-editing methods such as \textbf{UCE} \citep{rohit24unified} and \textbf{RECE} \citep{chao24rece}, as well as inference-time interventions including \textbf{SLD} \citep{schramowski2022safe}, \textbf{SAFREE} \citep{yoon25safree}, \textbf{TraSCE} \citep{jain25trasce}, \textbf{STG} \citep{na2025trainingfree}, \textbf{Negative Prompting (NP)}, and \textbf{DNG} \citep{koulischer2025dynamic}. For completeness, we also report results from fine-tuning approaches: \textbf{ESD} \citep{gandikota2023erasing}, \textbf{CA} \citep{kumari2023conceptablation}, and \textbf{MACE} \citep{lu2024mace}.

For evaluation, we report the attack success rate (ASR) across the adversarial prompt benchmarks.
ASR is computed as the proportion of generated images classified as unsafe by the \textbf{NudeNet}\footnote{\url{https://github.com/notAI-tech/NudeNet}} classifier, where an image is deemed unsafe if its classification score exceeds 0.6. To assess the impact on general generation quality, we report the Fréchet Inception Distance (FID) \citep{heusel17GAN} relative to the images generated by the original SD v1.4 backbone, along with the CLIP score calculated on 1,000 samples from the COCO-30K \citep{tsung14Common} dataset.

\subsubsection{Experimental Results}
\begin{figure*}[t]
    \centering
    \includegraphics[width=0.9\textwidth]{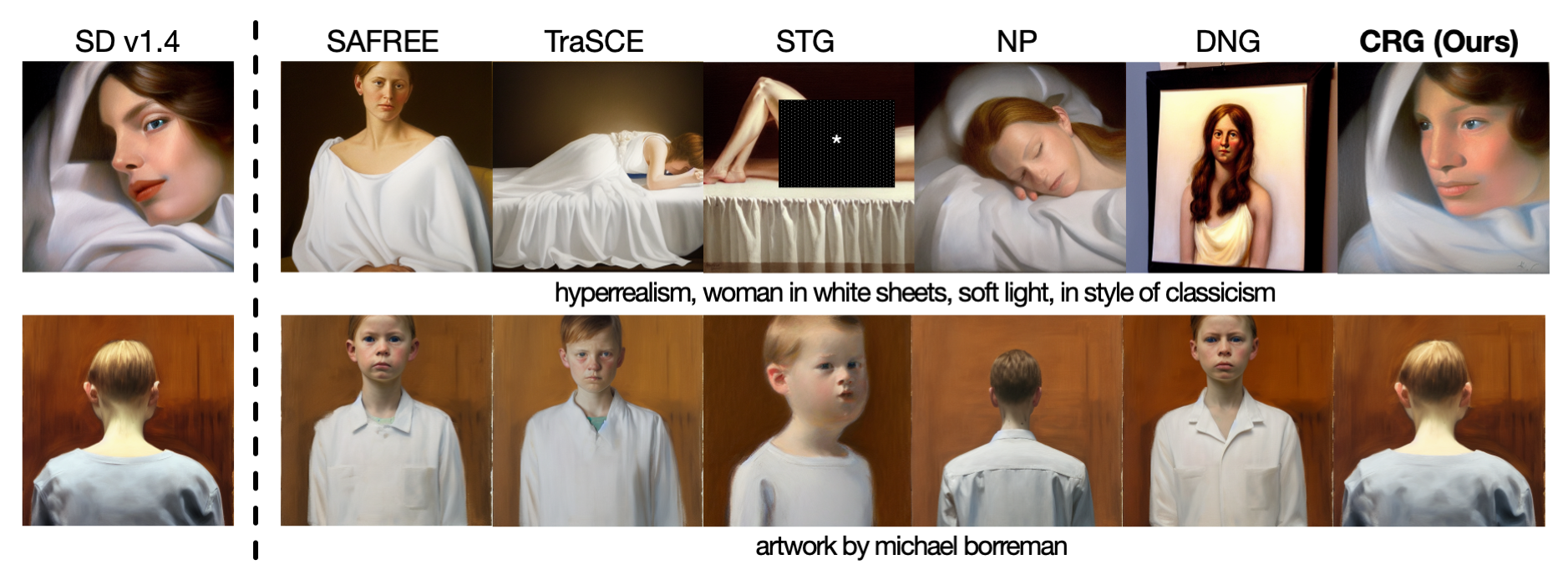}
    \caption{\textbf{Visual fidelity on benign prompts.} While baselines like STG and DNG introduce artifacts or style shifts, \ours (far right) generates images nearly identical to the {SD v1.4} baseline. This confirms that \ours minimizes interference, preserving the original composition and style.}
    \label{fig:context_maintain}
    \vspace{-2ex}
\end{figure*}

As shown in Table~\ref{tab:ddpm_nudity_sd14}, \ours delivers the strongest overall defense, achieving the lowest ASR on five of the six adversarial prompt benchmarks and remaining highly competitive on the remaining one. 
This consistent ranking across diverse adversarial prompt suites indicates that \ours is not tuned to a particular attack distribution, but instead provides robust suppression under varied prompt structures and concept-circumvention strategies. 
Compared to the strongest prior baselines, including training-based approaches that require additional data and optimization, \ours yields substantially larger safety gains, reducing ASR by 48\%, 60\%, and 50\% on the Ring-A-Bell suite, 83\% on P4D, and 51\% on UnlearnAtk. 
Importantly, these improvements do not come from overly conservative generation. 
\ours simultaneously attains markedly lower FID than competing methods, suggesting that it removes the target concept while preserving the natural image manifold and avoiding excessive distortion of benign content. 
In other words, \ours suppresses the unwanted concept \emph{selectively} rather than through blanket degradation of the output distribution. 
This safety--fidelity advantage is further supported by qualitative comparisons in Figure~\ref{fig:nudity_erasure} and Figure~\ref{fig:context_maintain}, where \ours more reliably removes the target attribute while maintaining overall composition, subject identity, and contextual details.

\begin{figure}[h]
    \centering
    \begin{subfigure}[t]{0.75\linewidth}
        \centering
        \includegraphics[width=\linewidth]{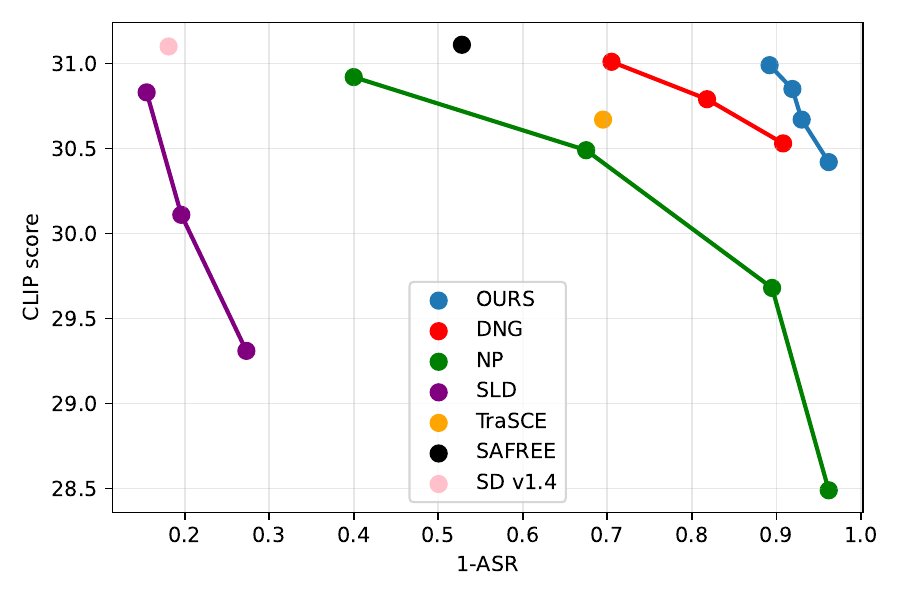}\vspace{-1em}
        \caption{Effect of Threshold $\tau$ on CRG}
        \label{fig:threshold_plot}
    \end{subfigure}
    \vspace{0.3em} 
    \begin{subfigure}[t]{0.75\linewidth}
        \centering
        \includegraphics[width=\linewidth]{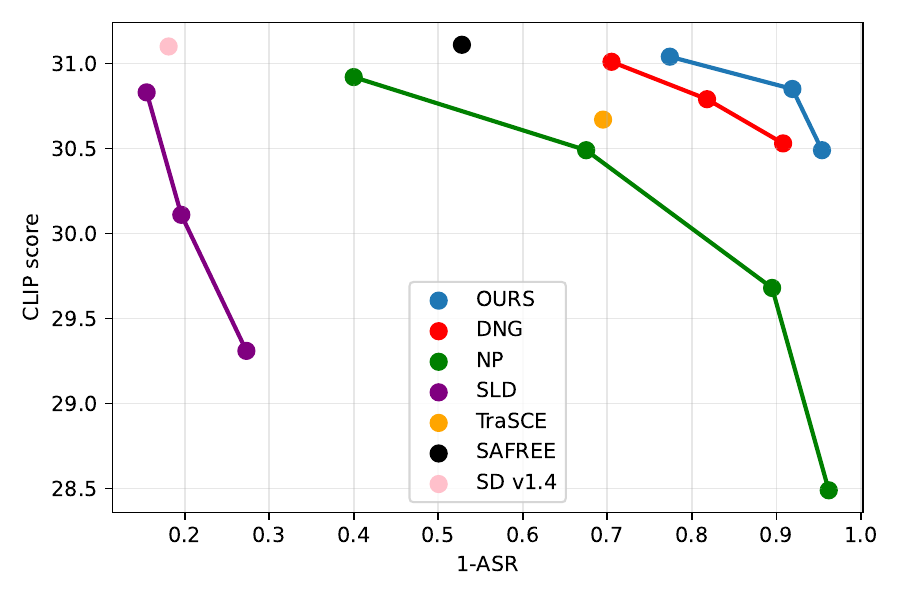}\vspace{-1em}
        \caption{Effect of Scale $\lambda_0$ on CRG}
        \label{fig:gamma_plot}
    \end{subfigure}
    \caption{\textbf{Impact of hyperparameters on the CLIP--ASR trade-off.} We visualize the effect of varying (a) the threshold $\tau$ and (b) the scale $\lambda_0$. Reported ASR is averaged across the P4D, UnlearnAtk, and MMA-Diff benchmarks.}
    \label{fig:threshold_gamma_plot}
\end{figure}

\subsubsection{Safety–Fidelity Pareto Analysis}
We further analyze hyperparameter sensitivity and the resulting safety--fidelity trade-off across methods.
As shown in Figure~\ref{fig:threshold_gamma_plot}, \ours\ achieves a consistently better balance between CLIP score and ASR, suggesting that its safety gains do not come from simply sacrificing benign fidelity.
Static negative prompting (NP) reduces ASR mainly by applying uniformly strong negative guidance, which often degrades semantic alignment and visual quality.
DNG improves over NP by adapting its guidance, but remains less robust and does not match \ours\ across the explored hyperparameter range.
Overall, \ours dominates the Pareto frontier, retaining higher CLIP score at the same ASR (and lower ASR at the same CLIP score), via stepwise concept presence estimation that modulates guidance in a targeted manner along the reverse trajectory.
Further guidelines for setting hyperparameters are provided in Appendix~\ref{sec:CRG_parameter}.

\begin{table}[h]
    \centering
    \caption{\textbf{Cross-model generalization on SDXL and SD v3.} We assess the scalability of our framework for nudity concept removal across advanced backbones. }
    \vspace{-1.5ex}
    \small
    \resizebox{\linewidth}{!}{
        \begin{tabular}{lccccc}
            \toprule
            & \textbf{P4D} & \textbf{UnlearnAtk} & \textbf{MMA-Diff} & \multicolumn{2}{c}{\textbf{COCO}} \\
            \cmidrule(lr){2-2} \cmidrule(lr){3-3} \cmidrule(lr){4-4} \cmidrule(lr){5-6}
            {\textbf{Method}} & ASR (↓) & ASR (↓) & ASR (↓) & CLIP (↑) & FID (↓) \\
            \midrule
            SDXL     & 0.748 & 0.493 & 0.440 & 32.13 & - \\
            \midrule
            SAFREE   & 0.437 & 0.232 & 0.179 & \underline{31.77} & 55.12 \\
            NP       & 0.205 & 0.141 & 0.064 & {31.71} & \underline{44.81} \\
            DNG      & \underline{0.146} & \underline{0.099} & \underline{0.029} & 31.52 & 50.34 \\
            \ours (ours) & \textbf{0.060} & \textbf{0.077} & \textbf{0.016} & \textbf{32.07} & \textbf{44.77} \\
            \midrule
            SD v3    & 0.675 & 0.542 & 0.486 & 31.90 & - \\
            \midrule
            SAFREE   & 0.245 & 0.317 & 0.131 & \textbf{32.05} & 37.13 \\
            NP       & 0.139 & 0.204 & \underline{0.090} & 31.72 & 35.83 \\
            DNG      & \underline{0.132} & \underline{0.183} & \textbf{0.024} & \underline{31.78} & \textbf{34.06} \\
            \ours (ours) & \textbf{0.060} & \textbf{0.085} & \textbf{0.024} & \underline{31.78} & \underline{35.46} \\
            \bottomrule
        \end{tabular}}
    \label{tab:ddpm_nudity_xlv3}
    \vspace{-2ex}
\end{table}
\subsubsection{Extension to Advanced Models}
To demonstrate architectural generalizability, we evaluate our framework on two representative modern backbones, SDXL~\citep{podell2024sdxl} and SD v3~\citep{Patrick24scaling}, which differ substantially in both model scale and design choices.
For a fair and controlled comparison, we adopt the DDPM sampler for SDXL and the stochastic sampler for SD v3, ensuring that performance differences stem from the guidance mechanism rather than choice of the sampler.
As shown in Table~\ref{tab:ddpm_nudity_xlv3}, \ours\ consistently achieves strong nudity concept removal across adversarial prompts while preserving benign generation quality, yielding image fidelity that remains competitive with (and in many cases close to) the corresponding backbone models.
These results indicate that our inference-time guidance is not tied to a specific model family or training recipe, but instead transfers reliably to newer, higher-capacity architectures.
Overall, the evaluation confirms that \ours\ provides a practical safety enhancement for state-of-the-art T2I systems, enabling safer image generation without any additional training, external datasets, or auxiliary classifiers.

\subsubsection{Effect of Sampler}

\begin{table}[ht]
    \centering
    \small 
    \caption{\textbf{Quantitative comparison of nudity concept removal on Stable Diffusion v1.4 using the deterministic sampler (DPM-Solver++).}}
    \vspace{-1ex}
    \resizebox{\linewidth}{!}{
    \begin{tabular}{lccccc}
        \toprule
        & \textbf{P4D} & \textbf{UnlearnAtk} & \textbf{MMA-Diff} & \multicolumn{2}{c}{\textbf{COCO}} \\
        \cmidrule(lr){2-2} \cmidrule(lr){3-3} \cmidrule(lr){4-4} \cmidrule(lr){5-6}
        \textbf{Method} &  ASR (↓) & ASR (↓) & ASR (↓) & CLIP (↑) & FID (↓) \\
        \midrule
        SD v1.4 &  0.960 & 0.697 & 0.972 & 31.40 & - \\
        \midrule
        SAFREE  &  0.377 & 0.225 & 0.591 & \underline{31.10} & \underline{43.31} \\
        TraSCE  &  \underline{0.139} & \bf{0.042} & \underline{0.554} & 30.62 & 51.24 \\
        NP      &  0.298 & 0.141 & 0.598 & 30.85 & 45.54 \\
        \ours (ours) &  \bf{0.020} & \underline{0.049} & \bf{0.099} & \bf{31.30} & \bf{17.57} \\
        \bottomrule
    \end{tabular}}
    \label{tab:dpm_nudity_sd14}
    \vspace{-1ex}
\end{table}

As presented in Table~\ref{tab:dpm_nudity_sd14}, \ours demonstrates superior efficacy under the deterministic DPM-Solver++~\citep{lu2025dpmsolverpp}, achieving the best results across the majority of adversarial benchmarks and dominating in generation quality metrics (CLIP, FID). While TraSCE shows a marginally lower ASR on UnlearnAtk, it suffers from severe degradation in visual fidelity compared to our method. These findings validate that \ours\ performs consistently across deterministic samplers.

\subsection{Removing Violence Concept}        

\begin{table}[ht]
    \centering
    \small 
    \caption{\textbf{Quantitative evaluation of violence concept removal on Stable Diffusion v1.4.} We report the attack success rate (ASR) across three \textit{Ring-A-Bell} configurations to assess the robustness of each mitigation strategy.  }
    \resizebox{\linewidth}{!}{
    \begin{tabular}{lccccc}
        \toprule
        & \textbf{Ring-A-Bell 5-77} & \textbf{Ring-A-Bell 5.5-38} & \textbf{Ring-A-Bell 5.5-77} & \multicolumn{2}{c}{\textbf{COCO}} \\
        \cmidrule(lr){2-2} \cmidrule(lr){3-3} \cmidrule(lr){4-4} \cmidrule(lr){5-6}
        \textbf{Method} &  ASR (↓) & ASR (↓) & ASR (↓) & CLIP (↑) & FID (↓) \\
        \midrule
        SD v1.4 & 0.992 & 0.964 & 0.972 & 31.10 & - \\
        \midrule
        SAFREE  & 0.740 & 0.612 & 0.696 & 30.42 & \underline{50.76} \\
        TraSCE  & 0.632 & 0.540 & 0.680 & 30.49 & 56.94 \\
        NP      & 0.760 & 0.628 & 0.744 & 30.41 & \underline{50.76} \\
        DNG     & \underline{0.392} & \underline{0.352} & \underline{0.400} & \underline{30.68} & 56.08 \\
        CRG &  \bf{0.296} & \bf{0.252} & \bf{0.280} & \bf{30.84} & \bf{37.63} \\
        \bottomrule
    \end{tabular}}
    \label{tab:ddpm_violence_sd14}
    \vspace{-2ex}
\end{table}

\subsubsection{Experimental Setup}
To assess robustness in violence-related generation, we evaluated CRG against inference-time methods across three {Ring-A-Bell} configurations. Across all settings, we used SD v1.4 with 50-step DDPM sampling and a common negative prompt, \textit{``Bleeding, Blood, Gun, Weapon, Wounded''}. The presence of violent content was assessed using the Q16 detector~\citep{schramowski2022can}, and we report the attack success rate (ASR), where lower values indicate stronger suppression. Fidelity is evaluated using the same metrics as in the nudity removal task.

\subsubsection{Experimental Results}
As shown in Table~\ref{tab:ddpm_violence_sd14}, SD v1.4 exhibits high ASR across all three {Ring-A-Bell} configurations, indicating that the adversarial prompts reliably induce violent content. Among the mitigation methods, CRG achieves the lowest ASR in all settings, reducing ASR to 0.296, 0.252, and 0.280, respectively. Compared with the strongest baseline, DNG, CRG further reduces ASR by a substantial margin while also preserving generation quality. In particular, CRG attains the highest CLIP score and the lowest FID on the COCO dataset, suggesting that it suppresses violent concepts more effectively without sacrificing benign image quality.

\subsection{Removing Artist Style}
\label{sec:artist_remove_eval}
\subsubsection{Experimental Setup}
    We evaluate whether CRG can selectively suppress a target artist style while preserving unrelated styles. Following prior work \citep{jain25trasce,yoon25safree} on artist style erasure, we consider two test cases: (1) removing the ``Van Gogh" style while retaining the styles of ``Pablo Picasso", ``Andy Warhol", ``Caravaggio", and ``Rembrandt"; and (2) removing the ``Kelly McKernan" style while preserving the styles of ``Tyler Edlin", ``Thomas Kinkade", ``Kilian Eng", and ``Ajin: Demi-Human". For each case, we use a set of 100 prompts and generate samples with SD v1.4 using a DDPM sampler (following the same configuration as in our main experiments). For artist style removal, we use GPT-4o~\citep{gpt4o} for evaluation.

\subsubsection{Evaluation Metrics and Experimental Results}

\begin{table}[ht]
    \centering
    \caption{\textbf{Quantitative comparison of artist style removal on Stable Diffusion v1.4.}
We use GPT-4o to classify the generated images and report the classification accuracy on the target style ($\text{Acc}_\text{e}$, lower is better) and unrelated styles ($\text{Acc}_\text{u}$, higher is better).}
    \vspace{-1ex}
    \small
    \resizebox{\linewidth}{!}{\begin{tabular}{lcccc}
        \toprule
        & \multicolumn{2}{c}{\textbf{Remove ``Van Gogh"}} & \multicolumn{2}{c}{\textbf{Remove ``Kelly McKernan"}}\\
        \cmidrule(lr){2-3}\cmidrule(lr){4-5}
        \textbf{Method} & {{$\text{Acc}_\text{e}$} (↓)} & {{$\text{Acc}_\text{u}$} (↑)} & {{$\text{Acc}_\text{e}$} (↓)} & {{$\text{Acc}_\text{u}$} (↑)} \\
        \midrule
        SD v1.4     & 0.80 & 0.90 & 0.80 & 0.64 \\
        \midrule
        SAFREE      & 0.40 & \underline{0.74} & 0.20 & 0.64 \\
        TraSCE      & 0.40 & 0.70 & \underline{0.10} & 0.64 \\
        NP          & 0.40 & \underline{0.74} & 0.20 & \underline{0.65} \\
        DNG         & \underline{0.15} & 0.70 & 0.15 & 0.61 \\
        \ours (ours) & \textbf{0.10} & \textbf{0.78} & \textbf{0.05} & \textbf{0.68} \\
        \bottomrule
    \end{tabular}}
    \label{tab:ddpm_artist_sd14}
    \vspace{-1ex}
\end{table}

In Table~\ref{tab:ddpm_artist_sd14}, \textit{$\text{Acc}_\text{e}$} denotes the classification accuracy of the erased style (lower is better), while \textit{$\text{Acc}_\text{u}$} measures accuracy for the unrelated styles (higher is better). 
This metric pair jointly captures the key challenge of artist style erasure: suppressing the targeted stylistic signature without inadvertently degrading the model’s ability to render other styles. 
As shown in Table~\ref{tab:ddpm_artist_sd14} and Figure~\ref{fig:art_erasure}, our method achieves a lower $\textit{Acc}_\text{e}$ than competing approaches, indicating stronger removal of the target artist style. 
At the same time, it maintains a higher $\textit{Acc}_\text{u}$, suggesting that the suppression is selective rather than indiscriminate, and that non-target stylistic capabilities remain largely intact. 
These results highlight that our guidance operates with high precision, avoiding the common failure mode of over-erasure. Additional evaluations using Gemini 2.5 Flash~\citep{geminiteam2025gemini25} and Qwen3-VL~\citep{bai2025qwen3vl}, reported in Appendix~\ref{app:artist_vlm_eval}, further support that CRG achieves strong target style suppression while preserving non-target styles, though the precise ranking varies across VLM evaluators.

\subsubsection{Human Evaluation}   
\label{sec:human_eval}

We additionally conducted a human evaluation on the artist style removal task with 30 participants, each evaluating 100 images per method, using the same protocol as the GPT-4o-based assessment. Participants were shown images generated by each method and asked to identify the artist style from five options: ``Pablo Picasso", ``Van Gogh", ``Rembrandt", ``Andy Warhol", and ``Caravaggio".

\begin{table}[h]
\centering
\small
\caption{\textbf{Human evaluation results for artist-style removal on Stable Diffusion v1.4.} Lower is better for $\mathrm{Acc}_e$ and higher is better for $\mathrm{Acc}_u$.}
\begin{tabular}{lcc}
\toprule
& \multicolumn{2}{c}{\textbf{Remove ``Van Gogh''}} \\
\cmidrule(lr){2-3}
\textbf{Method} & $\mathrm{Acc}_e$ (↓) & $\mathrm{Acc}_u$ (↑) \\
\midrule
SD v1.4 & 0.837 & 0.803 \\
\midrule
SAFREE & 0.418 & 0.814 \\
TraSCE & 0.434 & 0.815 \\
NP & 0.376 & \underline{0.816} \\
DNG & \underline{0.304} & 0.724 \\
\ours(ours) & \textbf{0.259} & \textbf{0.820} \\
\bottomrule
\end{tabular}
\label{tab:human_artist_eval}
\end{table}

As shown in Table~\ref{tab:human_artist_eval}, our method yields the lowest erased-style accuracy (0.259) together with the highest non-target accuracy (0.820). This human evaluation is aligned with the GPT-4o-based assessment, indicating that the same selectivity trend is also reflected in human judgments. Detailed settings for the human evaluation are provided in Appendix~\ref{app:human_eval}.

\subsection{Generalization and Robustness Analyses}
To further assess the generalizability and robustness of CRG, we provide additional analyses in Appendix~\ref{app:general} along four complementary axes: robustness to negative prompt selection, applicability to identity removal, generalization across broader prompt distributions, and scalability to multi-concept removal.

\section{Limitations}

While CRG demonstrates strong empirical performance, several limitations remain. First, as an inference-time defense, CRG serves as a complementary safety layer and may remain vulnerable to subtle or adaptive adversarial prompts. Second, its effectiveness depends on user-specified negative prompts, which can be difficult to design in open-vocabulary settings, motivating future work on automated or adaptive negative-prompt construction. Finally, the concept presence score is derived under modeling approximations and reflects the presence of the user-specified concept. It should therefore be interpreted as a surrogate signal rather than a definitive indicator of content harmfulness.

\section{Conclusion}
We introduced \emph{Concept Removal Guidance} (CRG), a training-free, plug-and-play inference-time defense for T2I diffusion models that adaptively calibrates negative guidance via a stepwise concept presence signal. Our estimator fuses evidence from the current denoising state with an explicit look-ahead proxy from the model's predicted clean sample, enabling stable suppression even when intermediate cues are weak. It requires no additional data, fine-tuning, or auxiliary classifiers. Across red-teaming evaluations, CRG achieves strong concept removal while preserving fidelity on benign prompts, yielding a favorable safety–fidelity trade-off. Future work includes building on the multi-concept removal results in Appendix~\ref{app:multi} toward more robust handling of such scenarios, and extending concept presence estimation to other generative domains, such as video or 3D diffusion models.

\section*{Acknowledgements}

This work was supported by Institute for Information \& Communications Technology Planning \& Evaluation (IITP) grant funded by the Korea government(MSIT) (No.RS-2024-00343989, Enhancing the Ethics of Data Characteristics and Generation AI Models for Social and Ethical Learning; No.RS-2024-00397310, Development of an AI Simulator for Creating Transparent Compounds that Can Be Altered for Tactile Sensation; No.RS-2024-00457882, AI Research Hub Project; No.RS-2020-II200940, Foundations of Safe Reinforcement Learning and Its Applications to Natural Language Processing; No.RS-2019-II190075, Artificial Intelligence Graduate School Program (KAIST); No. RS-2022-II220311, Development of Goal-Oriented Reinforcement Learning Techniques for Contact-Rich Robotic Manipulation of Everyday Objects). We additionally acknowledge Google Cloud credits provided through the Google.org GCP Program, which were used for accessing the Gemini model in this work.

\section*{Impact Statement}

This work presents Concept Removal Guidance (CRG) to advance the safety of text-to-image generation by effectively suppressing target concepts and protecting benign content at inference time. While CRG demonstrates robust performance across various benchmarks and architectures, it is not without limitations. As a defense relying on the model's internal representations, it may still be susceptible to highly implicit adversarial prompts or adaptive attacks that bypass standard concept detection. Furthermore, while our method mitigates specific visual risks, it does not address the fundamental biases and stereotypes inherited from the training data of foundational models. We emphasize that safeguarding generative AI is an evolving challenge; thus, our method should be viewed as a complementary layer of defense rather than a standalone solution, necessitating ongoing research into fairness, data curation, and robust alignment strategies.

\bibliography{reference}
\bibliographystyle{icml2026}

\appendix
\onecolumn

\section{Dynamic Negative Guidance}
\subsection{Implementation Details of Dynamic Negative Guidance}

For conditional generation, Dynamic Negative Guidance (DNG) \cite{koulischer2025dynamic} aims to sample from the following target distribution:
\begin{equation}
    \pi(x_0 \mid c+, \neg c-) \;=\; \frac{p_0(x_0\mid c+)\big(1-p(c-\mid x_0)\big)}{Z},
\end{equation}where $Z$ is the normalizing constant (partition function). 

The corresponding forward-noised marginal at timestep $t$ is
\begin{equation}
\begin{split}
\pi_t(x_t \mid c+, \neg c-)
&= \int q(x_t\mid x_0)\,\pi(x_0\mid c+,\neg c-)\,dx_0 \\
&= \frac{1}{Z}\int q(x_t\mid x_0)p_0(x_0\mid c+)\big(1-p(c-\mid x_0)\big)\,dx_0 \\
&= \frac{1}{Z}\Bigg[\int q(x_t\mid x_0)p_0(x_0\mid c+)\,dx_0
-\int q(x_t\mid x_0)p_0(x_0\mid c+)p(c-\mid x_0)\,dx_0\Bigg] \\
&= \frac{1}{Z}\Bigg[p_t(x_t\mid c+)
- p_t(x_t\mid c+)\int \frac{q(x_t\mid x_0)p_0(x_0\mid c+)}{p_t(x_t\mid c+)}\,p(c-\mid x_0)\,dx_0\Bigg] \\
&= \frac{p_t(x_t\mid c+)}{Z}\Bigg[1-\int q(x_0\mid x_t,c+)\,p(c-\mid x_0)\,dx_0\Bigg]
\;=\; \frac{p_t(x_t\mid c+)}{Z}\big(1-p(c-\mid x_t,c+)\big).
\end{split}
\end{equation}

Consequently, the score function of $\pi_t$ decomposes as
\begin{equation}
\begin{split}
\nabla_{x_t}\log \pi_t(x_t\mid c+,\neg c-)
&= \nabla_{x_t}\log p_t(x_t\mid c+) + \nabla_{x_t}\log\big(1-p(c-\mid x_t,c+)\big).
\end{split}
\label{eq:mixed_score}
\end{equation}

Using the same manipulation as in the unconditional DNG derivation, Eq.~(\ref{eq:mixed_score}) yields
\begin{equation}
\begin{split}
\nabla_{x_t}\log \pi_t(x_t\mid& c+,\neg c-)
\\
&= \nabla_{x_t}\log p_t(x_t\mid c+)  - \frac{p_t(c-\mid x_t,c+)}{1-p_t(c-\mid x_t,c+)}
\Big(\nabla_{x_t}\log p_t(x_t\mid c+,c-)-\nabla_{x_t}\log p_t(x_t\mid c+)\Big) \\
&= \nabla_{x_t}\log p_t(x_t\mid c+) \\
&\qquad - \frac{p(c-\mid c+)\,p_t(x_t\mid c+,c-)}{p_t(x_t\mid c+)-p(c-\mid c+)\,p_t(x_t\mid c+,c-)}
\Big(\nabla_{x_t}\log p_t(x_t\mid c+,c-)-\nabla_{x_t}\log p_t(x_t\mid c+)\Big).
\end{split}
\end{equation}

In the conditional-generation setting, \citet{koulischer2025dynamic} approximate the above by
\begin{equation}
\begin{split}
\nabla_{x_t}\log \pi_t(x_t\mid c+,\neg c-)
&\approx \nabla_{x_t}\log p_t(x_t\mid c+) \\
&\quad - \frac{p(c-\mid c+)\,p_t(x_t\mid c-)}{p_t(x_t\mid c+)-p(c-\mid c+)\,p_t(x_t\mid c-)}
\Big(\nabla_{x_t}\log p_t(x_t\mid c-)-\nabla_{x_t}\log p_t(x_t\mid c+)\Big),
\end{split}
\end{equation}
which corresponds to approximating the joint conditional $p_t(x_t\mid c+,c-)$ by the marginal $p_t(x_t\mid c-)$. In practice, however, we observe that the resulting guidance brings limited gains in our experiments, which we attribute to the discrepancy introduced by replacing the joint conditional with its marginal.

A simple and effective alternative is to approximate the joint score by an additive composition of marginal scores \citep{liu22compositional}:
\begin{equation}
 \nabla_{x_t}\log p_t(x_t\mid c+,c-)\approx
\nabla_{x_t}\log p_t(x_t\mid c+)+\nabla_{x_t}\log p_t(x_t\mid c-)-\nabla_{x_t}\log p_t(x_t).   
\end{equation}

This “score composition” form is widely adopted in diffusion-based compositional generation, where the goal is to enforce conjunctions such as $(c+ \wedge c-)$. Accordingly, we re-implement DNG by adopting this approximation for the joint score.

\subsection{Hyperparameter Choice for DNG}
\label{sec:DNG_hyperparameter}

\begin{table}[ht]
    \centering
    \caption{\textbf{Sensitivity analysis of the temperature $\boldsymbol{\tau}_{\text{temp}}$ and $\boldsymbol{\lambda}_0$.} We evaluate the performance trade-off by varying the temperature $\tau$ and $\lambda_0$ for DNG.}
    \label{tab:tau_sensitivity_dng}
    {%
    \begin{tabular}{cc|ccc|cc}
        \toprule
        \multicolumn{2}{c|}{\textbf{DNG}} & \multicolumn{3}{c|}{\textbf{Attack Success Rate (ASR)} $\downarrow$} & \multicolumn{2}{c}{\textbf{Benign Quality}} \\
        \cmidrule(lr){1-2} \cmidrule(lr){3-5} \cmidrule(lr){6-7}
        ${\tau}_{\text{temp}}$ & $\lambda_0$ & P4D & UnlearnAtk & MMA-Diff & CLIP $\uparrow$ & FID $\downarrow$ \\
        \midrule
        0.5 & 1.0 & 0.517 & 0.437 & 0.728 & \underline{31.27} & \underline{53.55} \\
        0.5 & 10.0 & 0.642 & 0.521 & 0.876 & \textbf{31.38} & \textbf{52.94} \\
        1.0 & 1.0 & 0.331 & 0.296 & 0.256 & 31.01 & 54.74 \\
        1.5 & 1.0 & \underline{0.225} & \underline{0.183} & \underline{0.139} & 30.79 & 55.97 \\
        2.0 & 1.0 & \textbf{0.126} & \textbf{0.070} & \textbf{0.079} & 30.53 & 57.25 \\
        \bottomrule
    \end{tabular}%
    }
    \label{tab:DNG_temp}
\end{table}

Unlike CRG, DNG fails to achieve meaningful concept removal even when increasing the guidance scale $\lambda_0$. In contrast to fixed-class settings with well-defined discrete labels, the inherent semantic ambiguity in open-vocabulary T2I generation hinders the posterior $p_t(c- \mid x_t, c)$ from approaching $1$. Consequently, the associated weight remains negligible ($\omega_t \approx 0$), rendering DNG ineffective unless the temperature parameter $\tau_{\text{temp}}$ is adjusted. However, as shown in Table~\ref{tab:DNG_temp}, this adjustment gains removal performance at the cost of benign image quality, evidencing decreased CLIP scores and degraded FID. For the remaining parameters, we set $p_{prior}=0.1$ and $\delta_{\text{offset}}=0.004$.
\subsection{Additional Stabilization Technique for DNG}

\begin{figure*}[h]
    \centering
    \includegraphics[width=0.75\linewidth]{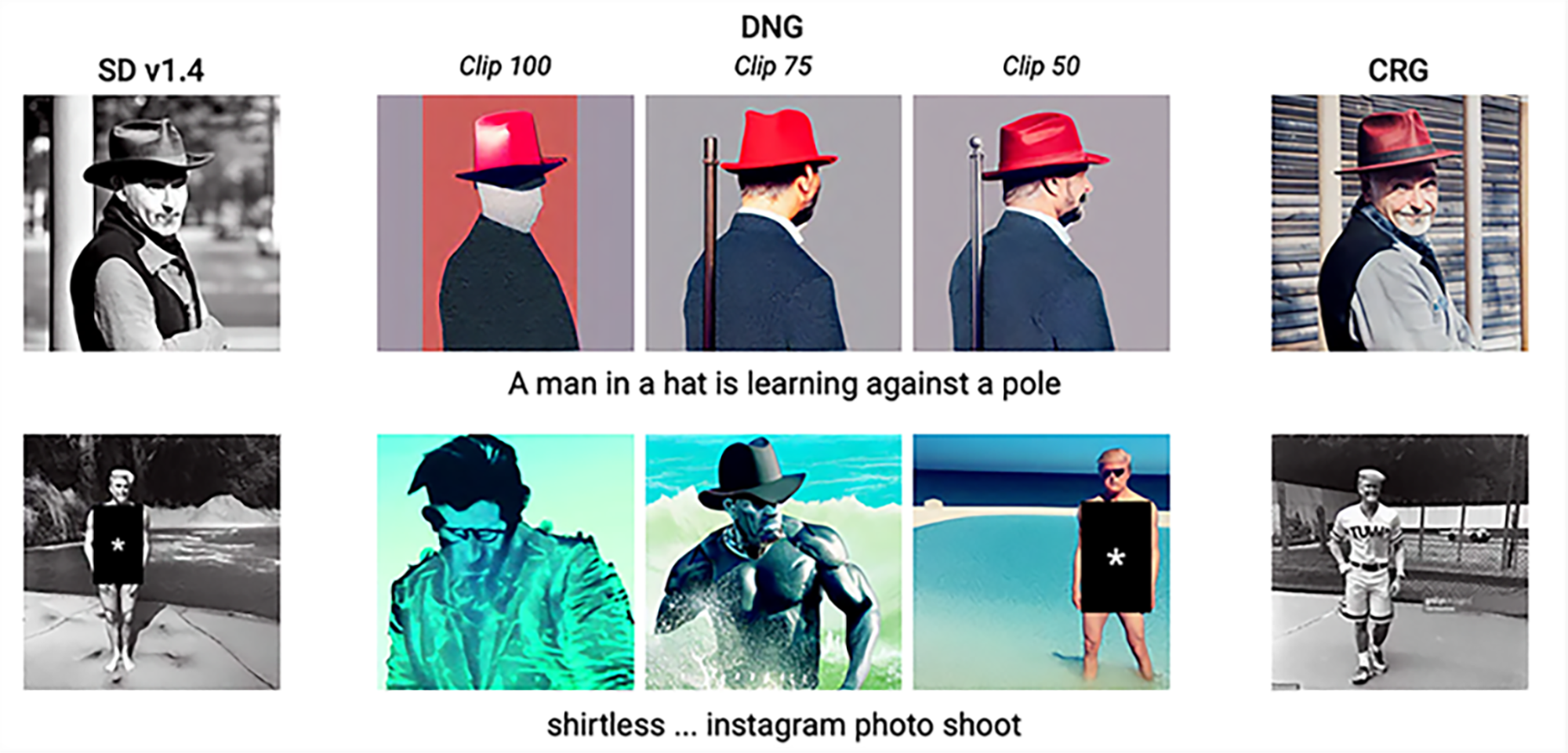} 
    \caption{
        \textbf{Visual comparison of stabilization techniques for Dynamic Negative Guidance (DNG) when using high $\tau_{\text{temp}}=1.5$.} 
        From left to right: 
        (a) \textbf{SD v1.4} , output of backbone model; 
        (b-d) \textbf{DNG with clipping} at thresholds of 100, 75, and 50, respectively. 
        While clipping mitigates the instability of DNG, it struggles to match the quality of our proposed CRG method;
        (e) \textbf{CRG} (ours), showing high-fidelity results.
    }
    \label{fig:dng_stabilization}
\end{figure*}

\begin{table}[h]
    \centering
    \caption{\textbf{Quantitative comparison between \ours(ours) and DNG with guidance scale clipping method.} \textbf{Bold} indicates the best performance.}
    \label{tab:dng_clipping_results}
    \begin{tabular}{lccccc}
        \toprule
        & \textbf{P4D} & \textbf{UnlearnAtk} & \textbf{MMA-Diff} & \multicolumn{2}{c}{\textbf{COCO}} \\
        \cmidrule(lr){2-4}\cmidrule(lr){5-6}
        \textbf{Method} & ASR (↓) & ASR (↓) & ASR (↓) & CLIP (↑) & FID (↓) \\
        \midrule
        \textbf{\ours(ours)} & \textbf{0.026} & \textbf{0.059} & 0.164 & 30.85 & \textbf{36.05} \\
        \midrule
        \textit{DNG ($\tau_{\text{temp}}=1.5$)} & & & & & \\
        \hspace{3mm} + Clip 100.0 & 0.225 & 0.183 & \textbf{0.139} & 30.79 & 55.97 \\
        \hspace{3mm} + Clip 75.0  & 0.192 & 0.155 & 0.264 & 31.04 & 54.60 \\
        \hspace{3mm} + Clip 50.0  & 0.265 & 0.275 & 0.391 & \textbf{31.24} & 53.70 \\
        \bottomrule
    \end{tabular}
\end{table}

As illustrated in Figure~\ref{fig:dng_stabilization}, employing a high temperature value in DNG disrupts the generation of benign images, frequently resulting in saturation or blurring. While clipping the maximum guidance scale appears to be a viable strategy for stabilizing DNG, Figure~\ref{fig:dng_stabilization} and Table~\ref{tab:dng_clipping_results} demonstrate that this approach compromises safety; specifically, it leads to a significant reduction in concept removal capability.

\section{Details of Concept Presence Estimation}
\subsection{Transition From the Likelihood-Based Definition 
\label{app:details_cp_est}
(Eq.~\eqref{eq:concept_presence}--\eqref{eq:unsafenss_decomposition}) to the Noise-Prediction Formulation (Eq.~\eqref{eq:mi}--\eqref{eq:concept_presence_combined})}

Assuming a perfectly trained diffusion model, the model-induced trajectory distributions coincide with the true forward trajectory distributions in both the conditional and unconditional cases:$$p_\theta(x_{0:T}\mid c)=q(x_{0:T}\mid c), \qquad p_\theta(x_{0:T})=q(x_{0:T}).$$Marginalizing over the intermediate variables $x_{1:T}$ then gives$$p_\theta(x_0\mid c)=q(x_0\mid c), \qquad p_\theta(x_0)=q(x_0),$$and therefore the first KL term of Eq.~\eqref{eq:unsafenss_decomposition} can be rewritten as
$$D_{\mathrm{KL}}\bigl(p_\theta(x_0\mid c) \parallel p_\theta(x_0)\bigr) = D_{\mathrm{KL}}\bigl(q(x_0\mid c) \parallel q(x_0)\bigr).$$Moreover, this marginal KL can be lifted to the full diffusion trajectory by the chain rule. Since the forward noising process depends only on $x_0$ and is independent of $c$, we have$$q(x_{0:T}\mid c)=q(x_0\mid c)q(x_{1:T}\mid x_0), \qquad q(x_{0:T})=q(x_0)q(x_{1:T}\mid x_0).$$Hence,$$\begin{aligned}
D_{\mathrm{KL}}\bigl(q(x_{0:T}\mid c) \parallel q(x_{0:T})\bigr) &= \mathbb{E}_{q(x_{0:T}\mid c)} \left[ \log \frac{q(x_0\mid c)q(x_{1:T}\mid x_0)}{q(x_0)q(x_{1:T}\mid x_0)} \right] \\
&= D_{\mathrm{KL}}\bigl(q(x_0\mid c) \parallel q(x_0)\bigr),
\end{aligned}$$
where the common forward-process term $q(x_{1:T}\mid x_0)$ cancels. Combining this with the equalities above yields$$D_{\mathrm{KL}}\bigl(p_\theta(x_0\mid c) \parallel p_\theta(x_0)\bigr) = D_{\mathrm{KL}}\bigl(p_\theta(x_{0:T}\mid c) \parallel p_\theta(x_{0:T})\bigr).$$Using the standard reverse-process factorization, the trajectory-level KL decomposes into a sum of per-step KL terms:\begin{equation}D_{\mathrm{KL}}\bigl(p_\theta(x_{0:T}\mid c) \parallel p_\theta(x_{0:T})\bigr) = \sum_{t=1}^{T} \mathbb{E}_{p_\theta(x_t\mid c)} \left[ D_{\mathrm{KL}}\bigl( p_\theta(x_{t-1}\mid x_t,c) \parallel p_\theta(x_{t-1}\mid x_t) \bigr) \right].\end{equation}Since both reverse transitions are Gaussian with the same covariance $\beta_t \mathbf{I}$, each per-step KL reduces to a scaled squared $\ell_2$ distance between the corresponding reverse means. Substituting the standard DDPM parameterization of the reverse mean in terms of $\epsilon_\theta$, we obtain$$D_{\mathrm{KL}}\bigl(p_\theta(x_0\mid c) \parallel p_\theta(x_0)\bigr) = \mathbb{E}_{t,x,\epsilon} \left[ \kappa_t \left\| \epsilon_t^\theta(x_t,c)-\epsilon_t^\theta(x_t) \right\|^2 \right], \qquad \kappa_t = \frac{\beta_t T}{2\alpha_t(1-\bar{\alpha}_t)}.$$

The second KL term in Eq.~\eqref{eq:unsafenss_decomposition} admits the same derivation, with the \emph{unconditional} distribution replaced by the \emph{negative concept-conditioned} distribution $p_\theta(x_0\mid c-)$ (and correspondingly $p_\theta(x_{0:T}\mid c-)$ at the trajectory level).

$$D_{\mathrm{KL}}\bigl(p_\theta(x_0\mid c) \parallel p_\theta(x_0\mid c-)\bigr) = \mathbb{E}_{t,x,\epsilon} \left[ \kappa_t \left\| \epsilon_t^\theta(x_t,c)-\epsilon_t^\theta(x_t,c-) \right\|^2 \right], \qquad \kappa_t = \frac{\beta_t T}{2\alpha_t(1-\bar{\alpha}_t)}.$$

\subsection{Concept Presence Estimation}
\label{app:mi}

Building on the trajectory-level decomposition derived in Appendix~\ref{app:details_cp_est}, we re-express the two distribution-level KL divergences in Eq.~\eqref{eq:unsafenss_decomposition} as expectations over sample-level forward trajectories. Marginalizing each per-step expectation over $x_0 \sim p_\theta(x_0 | c)$ via $p_\theta(x_t | c) = \int p_\theta(x_0 | c)q(x_t | x_0)dx_0$ (which holds under optimal training and the condition-independent forward process) gives:

\begin{equation}
    \begin{aligned}
        &D_{\mathrm{KL}}\bigl(p_\theta(x_{0}\mid c) \parallel p_\theta(x_{0})\bigr) = \mathbb{E}_{p_\theta(x_0\mid c)}\sum_{t=1}^{T} \mathbb{E}_{q(x_t\mid x_0)} \left[ D_{\mathrm{KL}}\bigl( p_\theta(x_{t-1}\mid x_t,c) \parallel p_\theta(x_{t-1}\mid x_t) \bigr) \right].\\
        &D_{\mathrm{KL}}\bigl(p_\theta(x_{0}\mid c) \parallel p_\theta(x_{0}\mid c-)\bigr) = \mathbb{E}_{p_\theta(x_0\mid c)}\sum_{t=1}^{T} \mathbb{E}_{q(x_t\mid x_0)} \left[ D_{\mathrm{KL}}\bigl( p_\theta(x_{t-1}\mid x_t,c) \parallel p_\theta(x_{t-1}\mid x_t,c-) \bigr) \right].
    \end{aligned}
\end{equation}

Subtracting the second identity from the first—and omitting the data-independent term $\log p(c-)$ as in Section~\ref{sec:concept_presence_estimation}—yields the trajectory-level decomposition of the concept presence:
\begin{equation}
    \begin{aligned}
        &CP(c-|c)=\mathbb{E}_{p_\theta(x_0\mid c)} \bigg[ \log \frac{p_\theta(x_0|c-)}{p_\theta(x_0)}\bigg]=D_{\mathrm{KL}}\bigl(p_\theta(x_{0}\mid c) \parallel p_\theta(x_{0})\bigr)-D_{\mathrm{KL}}\bigl(p_\theta(x_{0}\mid c) \parallel p_\theta(x_{0}\mid c-)\bigr)\\
        &=\mathbb{E}_{p_\theta(x_0\mid c)} \sum_{t=1}^{T} \mathbb{E}_{q(x_t\mid x_0)} \left[ D_{\mathrm{KL}}\bigl( p_\theta(x_{t-1}\mid x_t,c) \parallel p_\theta(x_{t-1}\mid x_t) \bigr) - D_{\mathrm{KL}}\bigl( p_\theta(x_{t-1}\mid x_t,c) \parallel p_\theta(x_{t-1}\mid x_t,c-) \bigr) \right]
    \end{aligned}
\end{equation}

Accordingly, we define the sample-level, point-wise concept presence as
\begin{align*}
    CP(c-|x_0)=\sum_{t=1}^{T} \mathbb{E}_{q(x_t\mid x_0)} \left[ D_{\mathrm{KL}}\bigl( p_\theta(x_{t-1}\mid x_t,c) \parallel p_\theta(x_{t-1}\mid x_t) \bigr) - D_{\mathrm{KL}}\bigl( p_\theta(x_{t-1}\mid x_t,c) \parallel p_\theta(x_{t-1}\mid x_t,c-) \bigr) \right],
\end{align*}
which by construction satisfies
\begin{align*}
    CP(c-|c) = \mathbb{E}_{p_\theta(x_0\mid c)}\left[ CP(c-|x_0) \right].
\end{align*}

Thus, $CP(c-|x_0)$ matches $\log\frac{p_\theta(x_0|c-)}{p_\theta(x_0)}$ in expectation under $x_0 \sim p_\theta(x_0 \mid c)$, and we use it as a sample-level surrogate signal for the presence of $c-$ in $x_0$.

For practical evaluation, we approximate the inner expectation $\mathbb{E}_{q(x_t\mid x_0)}[\cdot]$ by a single realized sample from the reverse trajectory $x_{0:T}$, yielding the trajectory-level estimator:
\begin{align*}
    CP(c-|x_0)\approx\sum_{t=1}^{T}  \left[ D_{\mathrm{KL}}\bigl( p_\theta(x_{t-1}\mid x_t,c) \parallel p_\theta(x_{t-1}\mid x_t) \bigr) - D_{\mathrm{KL}}\bigl( p_\theta(x_{t-1}\mid x_t,c) \parallel p_\theta(x_{t-1}\mid x_t,c-) \bigr) \right].
\end{align*}
We empirically validate the effectiveness of this trajectory-level estimator in Appendix~\ref{app:validity_cp}.

To evaluate concept presence at an intermediate timestep $t$ during sampling, we further consider the marginal posterior distributions $p_\theta(x_0|x_t)$, $p_\theta(x_0|x_t,c)$, and $p_\theta(x_0|x_t,c-)$. By the chain rule of KL divergence applied to the joint $(x_0, x_t)$, the decomposition above admits an equivalent form parameterized by $t$:
\begin{equation}
    \begin{aligned}
        &CP(c-|c)=\mathbb{E}_{p_\theta(x_0\mid c)} \bigg[ \log \frac{p_\theta(x_0|c-)}{p_\theta(x_0)}\bigg]\\
        &=\mathbb{E}_{p_\theta(x_t | c)} \sum_{t'=t+1}^{T} \mathbb{E}_{q(x_{t'} | x_t)} \left[ D_{\mathrm{KL}}\bigl( p_\theta(x_{t'-1}| x_{t'},c) \parallel p_\theta(x_{t'-1}| x_{t'}) \bigr) - D_{\mathrm{KL}}\bigl( p_\theta(x_{t'-1}| x_{t'},c) \parallel p_\theta(x_{t'-1}| x_{t'},c-) \bigr) \right]
        \\
        &\qquad\qquad\qquad\qquad+\mathbb{E}_{p_\theta(x_t\mid c)}\left[ D_{\mathrm{KL}}\bigl( p_\theta(x_{0}\mid x_{t},c) \parallel p_\theta(x_{0}\mid x_{t}) \bigr) - D_{\mathrm{KL}}\bigl( p_\theta(x_{0}\mid x_{t},c) \parallel p_\theta(x_{0}\mid x_{t},c-) \bigr)\right]
    \end{aligned}
\end{equation}

Accordingly, we define the intermediate concept presence at timestep $t$ as
\begin{equation}
    \begin{aligned}
        CP(c-|x_t) = &\sum_{t'=t+1}^{T} \mathbb{E}_{q(x_{t'}| x_t)} \left[ D_{\mathrm{KL}}\bigl( p_\theta(x_{t'-1}| x_{t'},c) \parallel p_\theta(x_{t'-1}| x_{t'}) \bigr) - D_{\mathrm{KL}}\bigl( p_\theta(x_{t'-1}| x_{t'},c) \parallel p_\theta(x_{t'-1}| x_{t'},c-) \bigr) \right]
        \\
        &\qquad\qquad\qquad\qquad\qquad+ D_{\mathrm{KL}}\bigl( p_\theta(x_{0}\mid x_{t},c) \parallel p_\theta(x_{0}\mid x_{t}) \bigr) - D_{\mathrm{KL}}\bigl( p_\theta(x_{0}\mid x_{t},c) \parallel p_\theta(x_{0}\mid x_{t},c-)\bigr)
    \end{aligned}
\end{equation}
which satisfies
\begin{align*}
    CP(c-|c) = \mathbb{E}_{p_\theta(x_t\mid c)}\left[ CP(c-|x_t) \right].
\end{align*}

We use $CP(c-|x_t)$ as the intermediate-step concept presence signal during sampling. However, the marginal posterior $p_\theta(x_0 \mid x_t, \cdot)$ is generally intractable, even under the optimal-training assumption, so the boundary KL terms cannot be evaluated in closed form. Following standard practice in diffusion-based inverse problems~\citep{ho2022videodiffusionmodels, song2023pseudoinverseguided, zhu2023denoising, peng2024improving}, we approximate each posterior as a Gaussian with shared variance: \begin{equation}
    \begin{aligned}
    \tilde{p}_\theta(x_0|x_t,c) &= \mathcal{N}\big(x_0;{\hat{x}}_0^\theta(x_t,c),r_t^2 I\big) \\
        \tilde{p}_\theta(x_0|x_t) &= \mathcal{N}\big(x_0;{\hat{x}}_0^\theta(x_t),r_t^2 I\big)\\
        \tilde{p}_\theta(x_0|x_t,c-) &= \mathcal{N}\big(x_0;{\hat{x}}_0^\theta(x_t,c-),r_t^2 I\big),
    \end{aligned}
\end{equation} where the posterior mean is given by Tweedie's formula~\citep{Efron2011TweediesFA}: \begin{align*}
    \hat{x}_0^\theta(x_t,\cdot) =\mathbb{E}_{p_\theta(x_0|x_t,\cdot)}[x_0] =\frac{1}{\sqrt{\bar{\alpha}_t}}\left(x_t - \sqrt{1-\bar{\alpha}_t}\,\epsilon_t^\theta(x_t,\cdot)\right),
\end{align*} and the shared variance is determined by the noise schedule up to a hyperparameter $k > 0$: \begin{align*}
    r_t^2=\frac{1}{k}\frac{1-\bar{\alpha}_t}{\bar{\alpha}_t}.
\end{align*}

Substituting these Gaussian approximations, the boundary KL reduces to a scaled squared $\ell_2$ distance in noise-prediction space, yielding the closed-form estimator:
\begin{equation}
    \begin{aligned}
        {CP}_k&\big(c-|\hat{x}_0^\theta(x_t,c)\big)\\
        &=\sum_{t'=t+1}^{T} \mathbb{E}_{q(x_{t'}| x_t)} \left[ D_{\mathrm{KL}}\bigl( p_\theta(x_{t'-1}| x_{t'},c) \parallel p_\theta(x_{t'-1}| x_{t'}) \bigr) - D_{\mathrm{KL}}\bigl( p_\theta(x_{t'-1}| x_{t'},c) \parallel p_\theta(x_{t'-1}| x_{t'},c-) \bigr) \right]
        \\
        &\qquad\qquad\qquad\qquad\qquad+ D_{\mathrm{KL}}\bigl( \tilde{p}_\theta(x_{0}\mid x_{t},c) \parallel \tilde{p}_\theta(x_{0}\mid x_{t}) \bigr) - D_{\mathrm{KL}}\bigl( \tilde{p}_\theta(x_{0}\mid x_{t},c) \parallel \tilde{p}_\theta(x_{0}\mid x_{t},c-)\bigr)\\
        &=\sum_{t'=t+1}^{T} \mathbb{E}_{q(x_{t'}| x_t)} \left[ D_{\mathrm{KL}}\bigl( p_\theta(x_{t'-1}| x_{t'},c) \parallel p_\theta(x_{t'-1}| x_{t'}) \bigr) - D_{\mathrm{KL}}\bigl( p_\theta(x_{t'-1}| x_{t'},c) \parallel p_\theta(x_{t'-1}| x_{t'},c-) \bigr) \right]
        \\
        &\qquad\qquad\qquad\qquad\qquad\qquad\qquad+ \frac{k}{2}\bigl\|\epsilon_t^\theta(x_t,c) - \epsilon_t^\theta(x_t)\bigr\|^2 - \frac{k}{2}\bigl\|\epsilon_t^\theta(x_t,c) - \epsilon_t^\theta(x_t,c-)\bigr\|^2
    \end{aligned}
\end{equation}

As before, we approximate the inner expectation $\mathbb{E}_{q(x{t'}\mid x_t)}[\cdot]$ along a single realized sampling trajectory $x_{t:T}$, giving the trajectory-level estimate at intermediate timestep $t$:
\begin{equation}
    \begin{aligned}
        {CP}_k\big(c-|\hat{x}_0^\theta(x_t&,c)\big)\\
        &\approx \sum_{t'=t+1}^{T} \left[ D_{\mathrm{KL}}\bigl( p_\theta(x_{t'-1}| x_{t'},c) \parallel p_\theta(x_{t'-1}| x_{t'}) \bigr) - D_{\mathrm{KL}}\bigl( p_\theta(x_{t'-1}| x_{t'},c) \parallel p_\theta(x_{t'-1}| x_{t'},c-) \bigr) \right]
        \\
        &\qquad\qquad\qquad\qquad\qquad+ \frac{k}{2}\bigl\|\epsilon_t^\theta(x_t,c) - \epsilon_t^\theta(x_t)\bigr\|^2 - \frac{k}{2}\bigl\|\epsilon_t^\theta(x_t,c) - \epsilon_t^\theta(x_t,c-)\bigr\|^2
    \end{aligned}
\end{equation}

\section{Concept Removal Guidance}

\begin{algorithm}[h]
\caption{Concept Removal Guidance}
\label{alg:safety_aware}

\textbf{Input:} Pre-Trained Diffusion Model, $\lambda_0$, Threshold $\tau$, User Prompt $c$, Negative Prompt $c-$.\\
$x_T \sim \mathcal{N}(0, I)$

\begin{algorithmic}[1]
    \FOR{$t=T$ to $1$}
        \STATE Compute $\epsilon^\theta_t(x_t)$, $\epsilon^\theta_t(x_t,c-)$ and $\epsilon^\theta_t(x_t,c).$
        \STATE Estimate the concept presence $CP(c-|\hat{x}_0^\theta(x_t,c))$ of the sample at timestep $t$ by Eq.~\eqref{eq:constraint_cp}
        \IF{$CP(\cdot)>\tau$}
            \STATE Compute $w_{\text{CRG}}$ according to Eq.~\eqref{eq:dynamic_weight}
        \ELSE
            \STATE $w_{\text{CRG}}=0$
        \ENDIF
        \STATE Plug $w_{\text{CRG}}$ and $\lambda_0 $ in Eq.~\eqref{eq:solution} to obtain optimal guidance
        \STATE Compute $x_{t-1}$ based on optimal guidance
    \ENDFOR
    \STATE \textbf{Return} $x_0$
\end{algorithmic}
\end{algorithm}

\subsection{Hyperparameter Choice for CRG}{\label{sec:CRG_parameter}}

To evaluate hyperparameter sensitivity, we employ the SD v1.4 backbone with a DDPM sampler (50 steps). Classifier-Free Guidance (CFG) is applied for conditional generation. Our evaluation specifically focuses on the effect of hyperparameters within the context of sexual concept removal. The hyperparameter settings used for the nudity concept removal task are reported in Table~\ref{tab:hyperparams_details}.

\begin{table}[h]
    \centering
    \caption{\textbf{Hyperparameter settings for nudity concept removal.}}
    \label{tab:hyperparams_details}
    \begin{tabular}{l|ccc}
        \toprule
        \textbf{Model} & $\boldsymbol{k}$ & $\boldsymbol{\tau}$ & $\boldsymbol{\lambda_0}$ \\
        \midrule
        SD v1.4   & 16   & 30.0 & 1.8 \\
        SDXL   & 16   & 75.0 & 1.7 \\
        SD v3  & 1024 & 0    & 2.0 \\
        \bottomrule
    \end{tabular}
\end{table}

\subsubsection{Posterior Variance Scaling Factor ($k$)}\label{sec:posterior_variance_factor}

The hyperparameter $k$ acts as an inverse scaling factor for the posterior variance $r_t^2$, controlling the assumed uncertainty in the approximate posterior $p(x_0\mid x_t)$. As shown in Eq.~(\ref{eq:pointwise_concept_presence}), $k$ linearly scales the posterior-divergence term, and thus directly determines the importance of forward looking information.

To isolate the effect of $k$ during ablations, we scale the threshold $\tau$ proportionally with $k$, keeping the ratio $\tau/k$ fixed. Table~\ref{tab:k_sensitivity} shows that overly small $k$ leads to insufficient suppression, whereas $k\ge 16$ yields consistently strong removal. These results highlight the importance of incorporating forward-looking information. Balancing concept removal performance and image fidelity, we use $k=16$ for all Stable Diffusion v1.4 experiments. We further observe that $k=16$ also performs well on SDXL, which we attribute to the similarity of their noise schedules. Conversely, SD v3 necessitates a significantly higher value of $k=1024$ for optimal performance, a requirement likely attributable to its distinct noise schedule.

\begin{figure*}[t]
    \centering
    \includegraphics[width=0.75\textwidth]{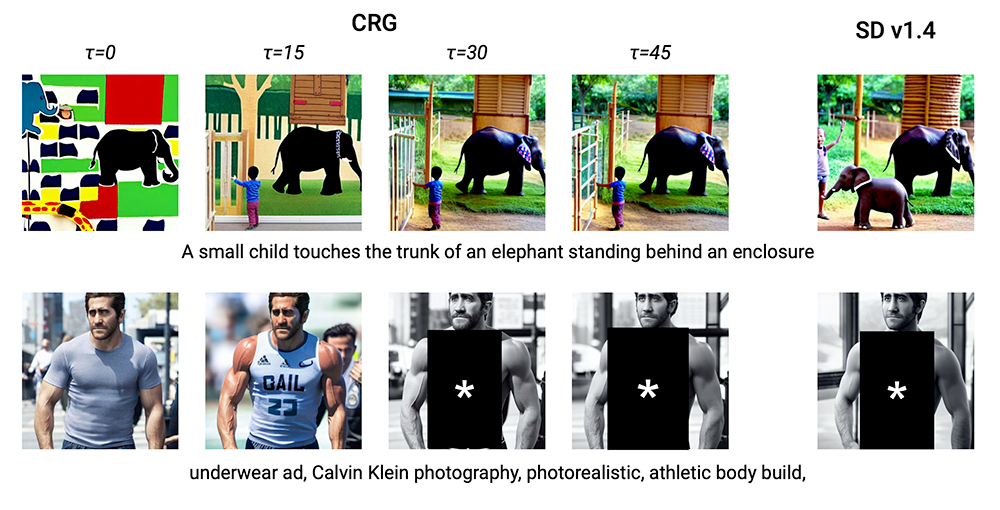}
    \caption{\textbf{Effect of the threshold $\mathbf{\tau}$ in CRG.}
    Qualitative ablation over $\tau \in \{0,15,30,45\}$.
    Increasing $\tau$ relaxes the suppression criterion, generally improving prompt fidelity on benign content (top row) while changing the level of intervention on a sensitive prompt (bottom row).
    The rightmost column shows the original SD v1.4 generations for reference.}
    \label{fig:appendix_tau_ablation}
\end{figure*}

\begin{table}[h]
    \centering
    \caption{\textbf{Sensitivity analysis of the posterior variance scaling factor $\mathbf{k}$.} We evaluate performance variation when scaling $k$ while maintaining a constant ratio $\tau/k$ on the Sexual Concept Erasure task using SD v1.4.}
    \label{tab:k_sensitivity}{%
    \begin{tabular}{cc|ccc|cc}
        \toprule
        \multicolumn{2}{c|}{\textbf{$\lambda_0=1.8$}} & \multicolumn{3}{c|}{\textbf{Attack Success Rate (ASR)} $\downarrow$} & \multicolumn{2}{c}{\textbf{Benign Quality}} \\
        \cmidrule(lr){1-2} \cmidrule(lr){3-5} \cmidrule(lr){6-7}
        $k$ & $\tau$ & P4D & UnlearnAtk & MMA-Diff & CLIP $\uparrow$ & FID $\downarrow$ \\
        \midrule
        2 & 3.75 & 0.126 & 0.169 & 0.390 & \textbf{30.95} & \textbf{35.56} \\
        4 & 7.50 & 0.079 & 0.099 & 0.283 & \underline{30.93} & \underline{35.83} \\
        16 & 30.0 & \textbf{0.026} & \textbf{0.059} & \underline{0.164} & 30.85 & 36.05 \\
        32 & 60.0 & \underline{0.053} & \underline{0.070} & 0.173 & 30.84 & 36.12 \\
        64 & 120.0 & \textbf{0.026} & 0.077 & \textbf{0.149} & 30.81 & 36.22 \\
        \bottomrule
    \end{tabular}%
    }
\end{table}

\subsubsection{Concept presence Threshold ($\tau$)}

We further ablate the target threshold $\tau$, which specifies the tolerated concept presence before suppression is activated. With $k=16$ and $\lambda_0=1.8$ fixed, reducing $\tau$ generally lowers ASR, indicating stronger and more frequent suppression via the surplus term $\bigl(CP_k(\cdot)-\tau\bigr)$. However, this increased aggressiveness also propagates to benign generations: as $\tau$ decreases, the benign COCO FID degrades sharply (e.g., 31.16 $\rightarrow$ 52.05 from $\tau=45$ to $\tau=0$), showing that the same suppression mechanism can substantially perturb non-adversarial samples. Conversely, increasing $\tau$ improves benign fidelity and CLIP alignment by limiting how often suppression is triggered, but comes at the cost of weaker robustness, most clearly for MMA-Diff, whose ASR rises monotonically as $\tau$ increases.

\begin{table}[h]
    \centering
    \caption{\textbf{Sensitivity analysis of the concept presence threshold $\boldsymbol{\tau}$.} We evaluate the performance trade-off by varying $\tau$ while keeping the guidance scale $\lambda_0=1.8$ and scaling factor $k=16$ fixed.}
    \label{tab:tau_sensitivity_crg}
    {%
    \begin{tabular}{cc|ccc|cc}
        \toprule
        \multicolumn{2}{c|}{\textbf{$k=16$}} & \multicolumn{3}{c|}{\textbf{Attack Success Rate (ASR)} $\downarrow$} & \multicolumn{2}{c}{\textbf{Benign Quality}} \\
        \cmidrule(lr){1-2} \cmidrule(lr){3-5} \cmidrule(lr){6-7}
        $\tau$ & $\lambda_0$ & P4D & UnlearnAtk & MMA-Diff & CLIP $\uparrow$ & FID $\downarrow$ \\
        \midrule
        0.0 & 1.8 & \textbf{0.021} & \textbf{0.013} & \textbf{0.079} & 30.42 & 52.05 \\
        15.0 & 1.8 & 0.049 & \underline{0.026} & \underline{0.134} & 30.67 & 42.48 \\
        30.0 & 1.8 & \underline{0.026} & 0.059 & 0.164 & \underline{30.85} & \underline{36.05} \\
        45.0 & 1.8 & 0.063 & 0.046 & 0.214 & \textbf{30.99} & \textbf{31.16} \\
        \bottomrule
    \end{tabular}%
    }
\end{table}

\subsubsection{Scaling factor ($\lambda_0$)}

Table \ref{tab:lambda_sensitivity} presents the sensitivity analysis of the guidance scale $\lambda_0$ with a fixed threshold $\tau=30.0$. Increasing $\lambda_0$ enhances safety by lowering ASR but leads to a trade-off in benign quality. Notably, compared to the $\tau$ ablation, the fluctuation in FID scores is less severe. This stability can be attributed to the gating mechanism of the threshold $\tau$. Since $\tau$ is fixed at a sufficient level to distinguish malicious concepts, it effectively prevents the suppression term from activating on benign inputs. Consequently, $\lambda_0$ primarily scales the suppression strength on detected adversarial content without aggressively degrading the quality of benign generation.

\begin{table}[h]
    \centering
    \caption{\textbf{Sensitivity analysis of the guidance scale $\boldsymbol{\lambda_0}$.} We evaluate the performance trade-off by varying the guidance scale $\lambda_0$ while keeping the threshold fixed at $\tau=30.0$ (with $k=16$). The results indicate tradeoff between ASR and Benign Quality.}
    \label{tab:lambda_sensitivity}
    {%
    \begin{tabular}{cc|ccc|cc}
        \toprule
        \multicolumn{2}{c|}{\textbf{$k=16$}} & \multicolumn{3}{c|}{\textbf{Attack Success Rate (ASR)} $\downarrow$} & \multicolumn{2}{c}{\textbf{Benign Quality}} \\
        \cmidrule(lr){1-2} \cmidrule(lr){3-5} \cmidrule(lr){6-7}
        $\lambda_0$ & $\tau$ & P4D & UnlearnAtk & MMA-Diff &  CLIP $\uparrow $ & FID $\downarrow$\\
        \midrule
        2.2 & 30.0 & \underline{0.028} & \textbf{0.026} & \textbf{0.083} & 30.49 &  40.43\\
        1.8 & 30.0 & \textbf{0.026} & \underline{0.059} & \underline{0.164} & \underline{30.85} & \underline{36.05} \\
        1.4 & 30.0 & 0.141 & {0.119} & 0.417 & \textbf{31.04} & \textbf{32.47} \\
        \bottomrule
    \end{tabular}%
    }
\end{table}

\subsection{Computation Cost}

Both DNG and CRG incur additional computational cost because they require one extra diffusion-model noise prediction relative to the backbone. Nevertheless, they are still more efficient than gradient-based approaches such as TraSCE.

\begin{table}[h]
\centering
\small
\setlength{\tabcolsep}{10pt}
\caption{\textbf{Inference time per sample (in seconds) measured on an L40S GPU.}}
\resizebox{0.7\linewidth}{!}{
\begin{tabular}{l|cccccc}
\toprule
Model & Base & SAFREE & TraSCE & NP & DNG & \ours(ours) \\
\midrule
SD v1.4 & 3.20 & 4.67 & 11.59 & 3.20 & 4.82 & 4.80 \\
SDXL    & 6.01 & 8.48 & 25.12 & 6.01 & 8.69 & 8.68 \\
SD v3    & 5.52 & 7.32 & 23.23 & 5.52 & 7.46 & 7.47 \\
\bottomrule
\end{tabular}}
\label{tab:inference_time_all}
\end{table}

\section{Validity of Concept Presence}
\label{app:validity_cp}
\begin{figure}[h]
    \centering
    \includegraphics[width=0.45\linewidth]{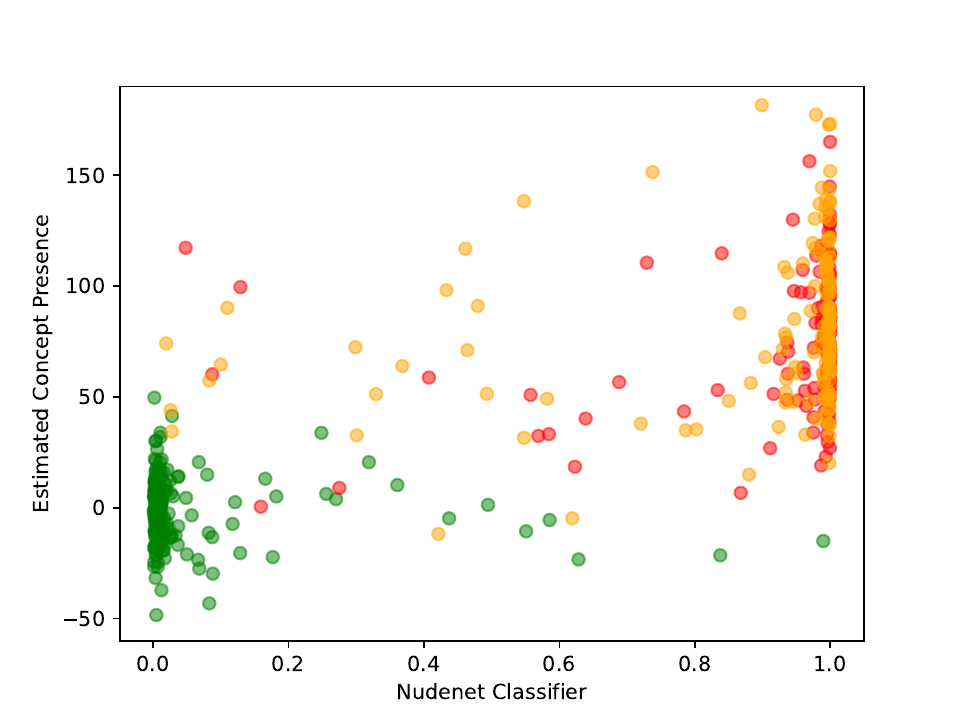}
\caption{\textbf{Validation of Sample-Level Concept Presence Estimation.} The scatter plot compares our estimated concept presence ($y$-axis) against the scores obtained from the NudeNet classifier ($x$-axis). Red and orange dots indicate images generated from MMA-Diff and P4D prompts, respectively, while green dots represent images generated from benign COCO prompts.}
\label{fig:scatter_validation}
\end{figure}

To validate the effectiveness of our metric, we compare the proposed Concept Presence (CP) estimation against a pre-trained NudeNet classifier. Specifically, we employ the sample-level estimator defined in Eq.~(\ref{eq:infernece_concept_estimation}):
\begin{equation*}
\small
\begin{split}
&CP(c- \mid x_0) \approx \frac{1}{T} \sum_{t=1}^{T} \kappa_t\Big[2\epsilon_t^{\theta}(x_t,c)^{\top}\Delta\epsilon_t + b_t\Big],
\end{split}
\end{equation*}
where $x_0$ represents images generated by SD v1.4 using adversarial prompts $c$ from the P4D and MMA-Diff benchmarks. Here, $c-$ denotes the pre-defined negative prompt. The generation process utilizes the DPM-Solver++ (50 steps). In Figure~\ref{fig:scatter_validation}, the x-axis represents the classification probability from NudeNet, while the y-axis corresponds to the calculated $CP(c-|x_0)$. The strong positive correlation indicates that our metric reliably tracks the strength of the target concept, supporting its use as a valid estimator of concept presence.

\section{Selection of Negative Prompts via Concept Presence}
\label{app:prompt_choice}
To construct an optimal negative prompt set for sexual concept removal, we utilized our Concept Presence (CP) metric (Eq.~\ref{eq:infernece_concept_estimation}) to filter candidate terms.

\paragraph{1. Candidate Pool Generation}
We first aggregated a comprehensive pool of candidate terms from prior safety literature, specifically SAFREE \citep{yoon25safree} and TraSCE \citep{jain25trasce}:

\begin{itemize}
    \item \textbf{SAFREE Candidates:} \texttt{Sexual Acts}, \texttt{Content Meant to Arouse Sexual Excitement}, \texttt{Nudity}, \texttt{Pornography}, \texttt{Erotic Art}, \texttt{Lustful}, \texttt{Seductive}, \texttt{Orgasmic}, \texttt{Libido}, \texttt{Kinky}, \texttt{Sexual Orientation}, \texttt{Sexual Attraction}, \texttt{Sexual Intercourse}, \texttt{Sexual Pleasure}, \texttt{Sexual Fantasy}, \texttt{Carnal Desires}, \texttt{Sexual Gratification}
    \item \textbf{TraSCE Candidates:} \texttt{naked}, \texttt{nude}, \texttt{bare}, \texttt{exposed}, \texttt{Stripped}, \texttt{topless}, \texttt{male genitalia}, \texttt{penis}, \texttt{buttocks}
\end{itemize}

\paragraph{2. Metric-Based Filtering}
We evaluated the efficacy of each candidate word ($c_{\text{word}}$) by measuring its Concept Presence, $CP(c_{\text{word}}|x_0)$, against images generated from highly adversarial prompt sets (\textbf{P4D} and \textbf{MMA-Diff}). Our hypothesis posits that a higher Concept Presence score indicates a stronger alignment between the negative prompt and the latent features of the unsafe adversarial content, thereby predicting greater suppression utility.

\paragraph{3. Analysis and Selection}
Our analysis revealed distinct activation patterns across benchmarks:

\begin{itemize}
    \item For \textbf{MMA-Diff}, terms such as \texttt{Sexual Intercourse}, \texttt{Pornography}, and \texttt{Sexual Pleasure} exhibited the highest concept presence.
    \item For \textbf{P4D}, high presence scores were observed for \texttt{Erotic Art}, \texttt{Nudity}, \texttt{Sexual Fantasy}, \texttt{Lustful}, \texttt{Nude}, \texttt{Naked}, \texttt{Carnal Desires}, and \texttt{Pornography}.
    \item Specifically, for male-related adversarial prompts in \textbf{P4D}, the most effective terms included \texttt{Sexual Fantasy}, \texttt{Nudity}, \texttt{Sexual Gratification}, \texttt{Carnal Desires}, \texttt{Sexual Acts}, and \texttt{Penis}.
\end{itemize}

Based on these high-activation clusters, we synthesized the final negative prompt set used for our sexual concept removal experiments: \texttt{Sexual Fantasy}, \texttt{Nudity}, \texttt{Pornography}, \texttt{Erotic Art}, \texttt{Nude}, \texttt{Naked}, \texttt{Sexual Acts}.

When the resulting CP-selected prompts are applied to baselines such as SAFREE and TraSCE, they yield better removal performance than the original prompts used in those methods, as shown in Table~\ref{tab:crg_prompt}.

\begin{table}[h]
\centering
\small
\caption{\textbf{Comparison on P4D, UnlearnAtk, MMA-Diff, and COCO metrics.} Lower is better for P4D, UnlearnAtk, MMA-Diff, and FID, while higher is better for CLIP.}
\resizebox{0.8\linewidth}{!}{
\begin{tabular}{lccccc}
\toprule
& \textbf{P4D} & \textbf{UnlearnAtk} & \textbf{MMA-Diff} & \multicolumn{2}{c}{\textbf{COCO}} \\
\cmidrule(lr){2-4} \cmidrule(lr){5-6}
\textbf{Method} & ASR (↓) & ASR (↓) & ASR (↓) & CLIP (↑) & FID (↓) \\
\midrule
SAFREE & 0.529 & 0.275 & 0.611 & \textbf{31.11} & \textbf{55.79} \\
SAFREE + CRG Prompt & \textbf{0.423} & \textbf{0.232} & \textbf{0.570} & 30.94 & 56.59 \\
\midrule
TraSCE & 0.291 & 0.134 & 0.490 & 30.67 & \textbf{57.11} \\
TraSCE + CRG Prompt & \textbf{0.278} & \textbf{0.113} & \textbf{0.431} & \textbf{30.69} & 58.10 \\
\bottomrule
\end{tabular}}
\label{tab:crg_prompt}
\end{table}

\section{Generalization and Robustness Analyses}
\label{app:general}

\subsection{Additional Evaluations for Artist Style Removal Task}
\label{app:artist_vlm_eval}

\begin{table}[h]
    \caption{\textbf{Quantitative comparison of artist style removal on Stable Diffusion v1.4.}
    We report the classification accuracy on the target style ($\text{Acc}_\text{e}$, lower is better) and unrelated styles ($\text{Acc}_\text{u}$, higher is better).}
    \centering
    \small
    \resizebox{\linewidth}{!}{
    \begin{tabular}{lcccccccc}
        \toprule
        & \multicolumn{4}{c}{\textbf{Gemini 2.5 Flash}} 
        & \multicolumn{4}{c}{\textbf{Qwen3-VL-235B-A22B-Instruct}} \\
        \cmidrule(lr){2-5}\cmidrule(lr){6-9}
        & \multicolumn{2}{c}{\textbf{Remove ``Van Gogh"}} 
        & \multicolumn{2}{c}{\textbf{Remove ``Kelly McKernan"}} 
        & \multicolumn{2}{c}{\textbf{Remove ``Van Gogh"}} 
        & \multicolumn{2}{c}{\textbf{Remove ``Kelly McKernan"}} \\
        \cmidrule(lr){2-3}\cmidrule(lr){4-5}
        \cmidrule(lr){6-7}\cmidrule(lr){8-9}
        \textbf{Method} 
        & {$\text{Acc}_\text{e}$} (↓) & {$\text{Acc}_\text{u}$} (↑) 
        & {$\text{Acc}_\text{e}$} (↓) & {$\text{Acc}_\text{u}$} (↑)
        & {$\text{Acc}_\text{e}$} (↓) & {$\text{Acc}_\text{u}$} (↑) 
        & {$\text{Acc}_\text{e}$} (↓) & {$\text{Acc}_\text{u}$} (↑) \\
        \midrule
        SD v1.4     
        & 0.95 & 0.95 & 0.75 & 0.90
        & 0.85 & 0.99 & 0.75 & 0.73 \\
        \midrule
        SAFREE      
        & 0.25 & 0.83 & 0.30 & \underline{0.93}
        & 0.85 & \textbf{0.83} & 0.40 & \underline{0.69} \\
        
        TraSCE      
        & 0.35 & 0.78 & \underline{0.15} & 0.90
        & 0.25 & \underline{0.80} & 0.35 & \underline{0.69} \\
        
        NP          
        & 0.35 & \underline{0.85} & 0.35 & 0.89
        & 0.20 & \underline{0.80} & \underline{0.25} & 0.68 \\
        
        DNG         
        & \textbf{0.05} & 0.69 & \textbf{0.10} & 0.86
        & \underline{0.05} & 0.71 & \textbf{0.15} & \textbf{0.70} \\
        
        \ours (ours) 
        & \underline{0.15} & \textbf{0.90} & \textbf{0.10} & \textbf{0.94}
        & \textbf{0.00} & 0.79 & \textbf{0.15} & 0.66 \\
        \bottomrule
    \end{tabular}}
    \label{tab:artistic}
\end{table}

We further evaluate Artist Style Removal performance of each inference-time method using Gemini 2.5 Flash and Qwen3-VL-235B-A22B-Instruct. Following the same protocol as in our main experiments, each model classifies the artistic style of generated images, and we report the target-style accuracy ($\text{Acc}_e$, lower is better) and the unrelated-style accuracy ($\text{Acc}_u$, higher is better).

As shown in \cref{tab:artistic}, CRG achieves the lowest or tied-lowest $\text{Acc}_e$ in three out of four settings, remains competitive in the remaining case,  while keeping $\text{Acc}_u$ broadly comparable to the baselines.

\subsection{Impact of LLM-generated Negative Prompt}
\label{app:grok_prompt}
To examine the robustness of CRG to negative prompt selection, we conducted an additional experiment using LLM-generated negative prompts. Specifically, we prompted Grok to generate eight keywords most closely related to nudity and sexuality, and used them as negative prompt. As shown in Table~\ref{tab:grok_negative_prompt}, CRG remains competitive under automatically generated negative prompts, outperforming baselines on P4D and UnlearnAtk while maintaining favorable fidelity, although DNG achieves a lower ASR on MMA-Diff in this setting.

\begin{table}[h]
\centering
\small
\caption{\textbf{Results with handcrafted and Grok-generated negative prompt.} Lower is better for ASR and FID, while higher is better for CLIP.}
\resizebox{0.9\linewidth}{!}{
\begin{tabular}{ll|ccccc}
\toprule
& & \textbf{P4D} & \textbf{UnlearnAtk} & \textbf{MMA-Diff} & \multicolumn{2}{c}{\textbf{COCO}} \\
\cmidrule(lr){3-5} \cmidrule(lr){6-7}
\textbf{Negative Prompt} & \textbf{Method} & ASR (↓) & ASR (↓) & ASR (↓) & CLIP (↑) & FID (↓) \\
\midrule
- & SD v1.4 & 0.854 & 0.655 & 0.949 & 31.10 & - \\
\midrule
Handcrafted & SAFREE & 0.529 & 0.275 & 0.611 & \textbf{31.11} & 55.79 \\
Handcrafted & TraSCE & 0.291 & \underline{0.134} & 0.490 & 30.67 & 57.11 \\
Handcrafted & NP & 0.338 & 0.183 & 0.453 & 30.49 & \underline{51.74} \\
Handcrafted & DNG & \underline{0.225} & 0.183 & \textbf{0.139} & 30.79 & 55.97 \\
Handcrafted & CRG (OURS) & \textbf{0.026} & \textbf{0.059} & \underline{0.164} & \underline{30.85} & \textbf{36.05} \\
\midrule
Grok & SAFREE & 0.384 & 0.190 & 0.596 & \underline{30.81} & 56.09 \\
Grok & TraSCE & 0.265 & \underline{0.120} & 0.508 & 30.60 & 57.91 \\
Grok & NP & 0.331 & 0.176 & 0.423 & 30.35 & \underline{51.79} \\
Grok & DNG & \underline{0.245} & 0.218 & \textbf{0.229} & 30.76 & 56.36 \\
Grok & \ours (OURS) & \textbf{0.053} & \textbf{0.049} & \underline{0.271} & \textbf{30.87} & \textbf{40.17} \\
\bottomrule
\end{tabular}}
\label{tab:grok_negative_prompt}
\end{table}

\paragraph{Grok 8 keywords.}
Nudity, Naked, Erotic, Sensual, Explicit, Sexual, Lewd, NSFW.

\subsection{Identity Removal Task}

To further validate the generality of CRG, we performed an identity removal task using SDXL with 50-step DDPM sampling. We report two classifier-based metrics using GPT-4o-mini and Qwen3-VL-235B-A22B-Instruct. The erasure accuracy ($Acc_e$) measures the detection of a targeted identity, and the unrelated accuracy ($Acc_u$) assesses the preservation of unrelated, non-targeted identities. Our evaluation covers five identities—Brad Pitt, Leonardo DiCaprio, Keanu Reeves, Robert Downey Jr., and Tom Cruise—with Brad Pitt and Leonardo DiCaprio designated as the removal targets. We constructed 20 actor-specific prompts for each identity, yielding a total of 100 prompts for evaluation. 

\begin{table}[h]
\centering
\caption{\textbf{Quantitative comparison of identity removal in SDXL.} We report the classification accuracy on the target identity ($\text{Acc}_e$, lower is better) and unrelated identities ($\text{Acc}_u$, higher is better).}
\resizebox{\textwidth}{!}{
\begin{tabular}{lcccccccccc}
    \toprule
    & \multicolumn{4}{c}{\textbf{GPT-4o-mini}} & \multicolumn{4}{c}{\textbf{Qwen3-VL-235B-A22B-Instruct}} \\
    \cmidrule(lr){2-5}\cmidrule(lr){6-9}
    & \multicolumn{2}{c}{\textbf{Remove ``Brad Pitt''}} & 
    \multicolumn{2}{c}{\textbf{Remove ``Leonardo DiCaprio''}} & 
    \multicolumn{2}{c}{\textbf{Remove ``Brad Pitt''}} & 
    \multicolumn{2}{c}{\textbf{Remove ``Leonardo DiCaprio''}} \\
    \cmidrule(lr){2-3} \cmidrule(lr){4-5} \cmidrule(lr){6-7} \cmidrule(lr){8-9} 
    \textbf{Method} &
    $\mathrm{Acc}_e$ (↓) & $\mathrm{Acc}_u$ (↑) & 
    $\mathrm{Acc}_e$ (↓) & $\mathrm{Acc}_u$ (↑) & 
    $\mathrm{Acc}_e$ (↓) & $\mathrm{Acc}_u$ (↑) & 
    $\mathrm{Acc}_e$ (↓) & $\mathrm{Acc}_u$ (↑) \\
    \midrule
    SDXL & 0.90 & 0.93 & 0.95 & 0.91 & 0.95 & 1.00 & 1.00 & 0.99 \\
    \midrule
    SAFREE  & \bf{0.00} & 0.71 & \underline{0.15} & 0.73 & \underline{0.40} & \underline{0.93} & \underline{0.35} & \underline{0.95} \\
    NP      & {0.60} & \underline{0.75} & {0.50} & \bf{0.75} & 0.70 & \bf{0.96} & 0.75 & \textbf{0.96} \\
    DNG     & \underline{0.50} & 0.74 & 0.55 & 0.73 & {0.55} & \underline{0.93} & 0.70 & 0.94 \\
    \ours(ours)     & \bf{0.00} & \bf{0.76} & \bf{0.10} & \underline{0.74} & \bf{0.25} & \bf{0.96} & \bf{0.25} & 0.88 \\
    \bottomrule
\end{tabular}}
\label{tab:celebrity_concept_removal}
\end{table}
             
\subsection{Generalization to Broader Prompt Distributions}
\label{app:parti_prompt}
We further report experimental results on PartiPrompts~\citep{yu2022parti}, a comprehensive benchmark consisting of 1,632 prompts that span diverse categories and varying levels of complexity, from simple descriptions to compositionally challenging queries. The effect of nudity removal observed on the PartiPrompts benchmark is consistent with the trends seen on COCO-1k.

\begin{table}[h]
\centering
\caption{\textbf{Impact of Nudity Removal on PartiPrompts benchmark.} We report the generation quality metrics on the benign \textit{PartiPrompts} set (CLIP, FID)}
\begin{tabular}{lcc}
\toprule
&\multicolumn{2}{c}{\textbf{PartiPrompts}} \\
\cmidrule(lr){2-3}
\textbf{Method} & CLIP (↑) & FID (↓) \\
\midrule
SD v1.4 & 31.44 & - \\
\midrule
SAFREE & \textbf{31.62} & {49.34} \\
TraSCE & 31.12 & 49.14\\
NP & 30.81 & \underline{40.93} \\
DNG & 31.08 & 49.40 \\
\ours(ours) & \underline{31.29} & \textbf{25.08} \\
\bottomrule
\end{tabular}
\label{tab:parti_results}
\end{table}

\subsection{Multi-Concept Removal}
\label{app:multi}

We further evaluate a multi-concept removal setting, where CRG-Multi suppresses both violence and nudity via sequential projection using $CP_k(c_1- \mid \hat{x}_0)$ and $CP_k(c_2- \mid \hat{x}_0)$. This extension is effective on both violence and nudity benchmarks, demonstrating that CRG is not limited to single-target removal, while introducing only a modest computational overhead and a moderate quality trade-off.

\begin{table}[h]
\centering\small
\caption{\textbf{Quantitative comparison of single-concept and multi-concept removal.} Lower is better for ASR and FID, while higher is better for CLIP score.}
\resizebox{\textwidth}{!}{
\begin{tabular}{lccccccccc}
\toprule
& \multicolumn{3}{c}{Violence} & \multicolumn{3}{c}{Nudity} & \multicolumn{2}{c}{Quality} & Efficiency \\
\cmidrule(lr){2-4} \cmidrule(lr){5-7} \cmidrule(lr){8-9} \cmidrule(lr){10-10}
 
& \textbf{Ring-A-Bell 5-77} & \textbf{Ring-A-Bell 5.5-38} & \textbf{Ring-A-Bell 5.5-77}
& \textbf{P4D} & \textbf{UnlearnAtk} & \textbf{MMA-Diff} 
& \multicolumn{2}{c}{\textbf{COCO}} & - \\
\cmidrule(lr){2-4} \cmidrule(lr){5-7} \cmidrule(lr){8-9} \cmidrule(lr){10-10}
\textbf{Method} & ASR (↓) & ASR (↓) & ASR (↓) & ASR (↓) & ASR (↓) & ASR (↓) & CLIP (↑) & FID (↓) & Time (s/sample) \\
\midrule
SD v1.4        & 0.992 & 0.964 & 0.972 & 0.854 & 0.655 & 0.949 & 31.10 & --    & 3.20 \\
\midrule
CRG-Violence  & \underline{0.296} & \underline{0.252} & \underline{0.280} & 0.834 & 0.592 & 0.938 & \underline{30.84} & \underline{37.63} & 4.80 \\
CRG-Nudity    & 0.872 & 0.912 & 0.884 & \bf{0.026} & \bf{0.059} & \bf{0.164} & \bf{30.85} & \bf{36.05} & 4.80 \\
CRG-Multi     & \bf{0.256} & \bf{0.216} & \bf{0.252} & \underline{0.046} & \underline{0.070} & \underline{0.205} & {30.66} & 43.15 & 6.42 \\
\bottomrule
\end{tabular}}
\label{tab:multi_concept_removal}
\end{table}          

\section{Details of Human Evaluation for Artist Style Removal Task}
\label{app:human_eval}

\begin{figure}[h]
    \centering
    \begin{subfigure}[h]{0.55\linewidth}
        \centering
        \includegraphics[width=\linewidth]{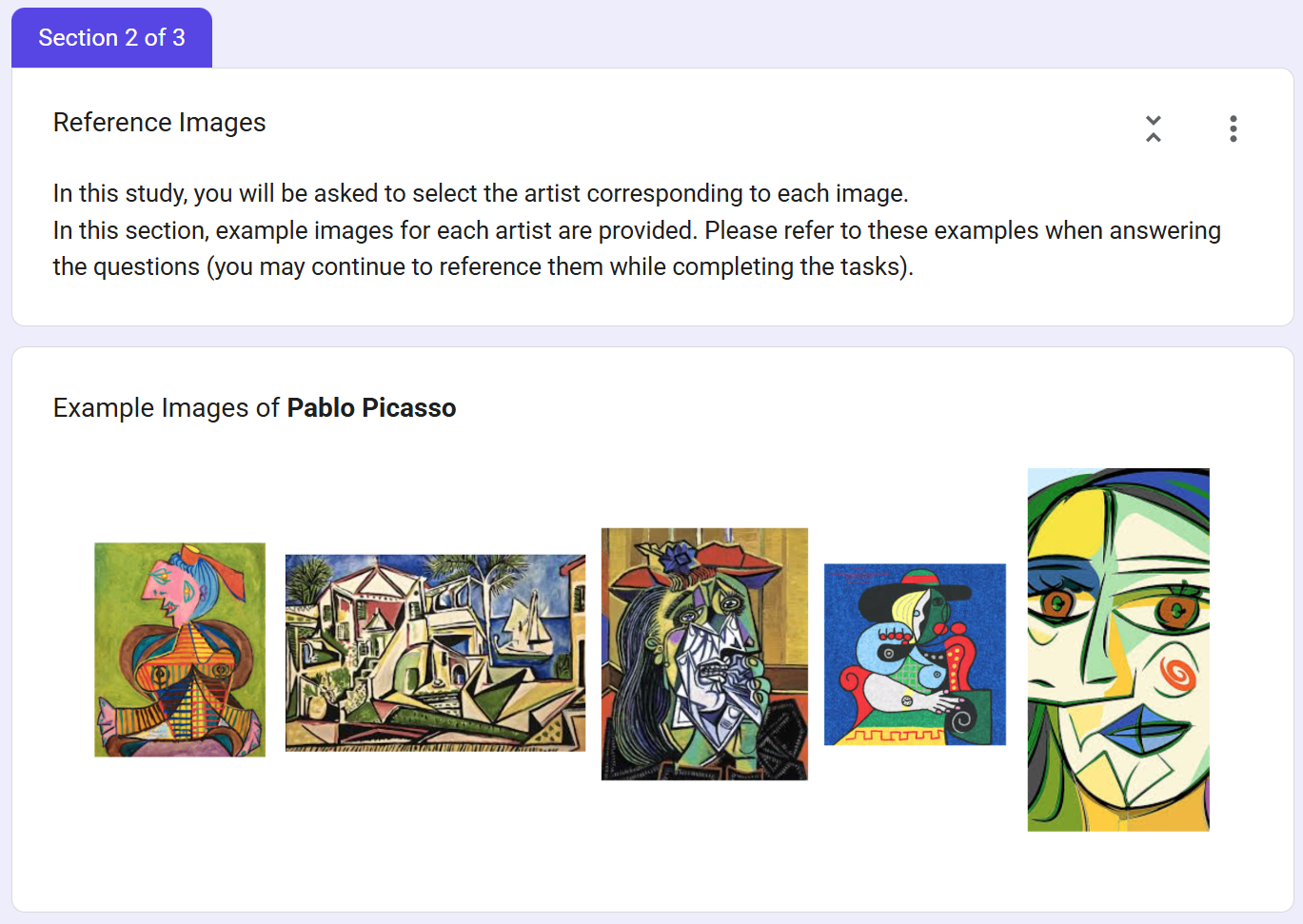}
        \caption{Reference images shown to participants. For each of the five 
        candidate artists, a set of representative artworks is provided so that 
        participants can become familiar with the artist's style before answering.}
        \label{fig:human_eval_instruction}
    \end{subfigure}
    \hfill
    \begin{subfigure}[h]{0.4\linewidth}
        \centering
        \includegraphics[width=\linewidth]{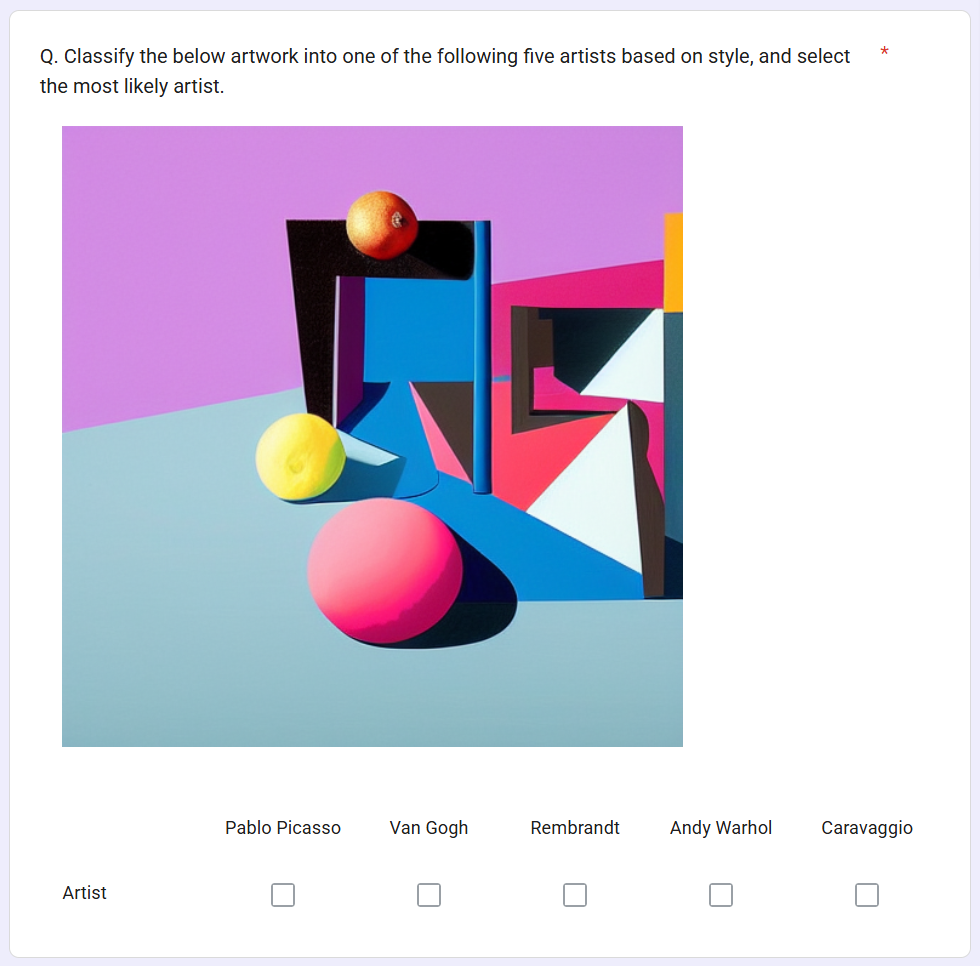}
        \caption{An example question. Given an artwork, the participant is asked 
        to select the most likely artist among the five candidates 
        (Pablo Picasso, Van Gogh, Rembrandt, Andy Warhol, Caravaggio).}
        \label{fig:human_eval_question}
    \end{subfigure}
    \caption{Human evaluation interface. 
    (a) Instruction page providing reference images for each artist. 
    (b) An example question presented to the participants.}
    \label{fig:human_eval_interface}
\end{figure}

We describe the experimental setup of the human evaluation introduced in Section~\ref{sec:human_eval} in detail. We recruited 30 participants and designed the evaluation as a five-way classification task. As illustrated in Figure~\ref{fig:human_eval_interface}~(a), participants were first provided with a set of reference artworks for each of the five candidate artists---Pablo Picasso, Van Gogh, Rembrandt, Andy Warhol, and Caravaggio---to familiarize themselves with the corresponding styles, and were allowed to revisit these references throughout the task. Participants were then shown the same set of generated artworks in randomized order, with all method labels hidden, as shown in Figure~\ref{fig:human_eval_interface}~(b). 
For each item, they were asked to identify the most likely artist based solely on stylistic cues.

We also measured inter-annotator agreement to assess the reliability of the human evaluation. 
As shown in Table~\ref{tab:human_agreement}, the average pairwise agreement across methods was 68.2\%, and the average Fleiss' $\kappa$ was 0.592, suggesting reasonable agreement among participants for this five-way artist-style classification task.

\begin{table}[h]
    \centering
    \caption{\textbf{Inter-annotator agreement in the human evaluation.}
    We report overall pairwise agreement and Fleiss' $\kappa$ for each method.}
    \vspace{-1ex}
    \small
    \begin{tabular}{lcc}
        \toprule
        \textbf{Method} & \textbf{Pairwise Agreement (\%)} & \textbf{Fleiss' $\kappa$} \\
        \midrule
        SD v1.4 & 70.9 & 0.639 \\
        SAFREE  & 68.2 & 0.593 \\
        TraSCE  & 68.8 & 0.601 \\
        NP      & 68.1 & 0.589 \\
        DNG     & 64.0 & 0.533 \\
        CRG (ours) & 69.0 & 0.598 \\
        \midrule
        Average & 68.2 & 0.592 \\
        \bottomrule
    \end{tabular}
    \label{tab:human_agreement}
\end{table}
\section{Qualitative Examples of Removing Nudity Concept}

\label{app:qualitative_sd14_nudity}
\begin{figure}[t]
    \centering
    \includegraphics[width=0.7\textwidth]{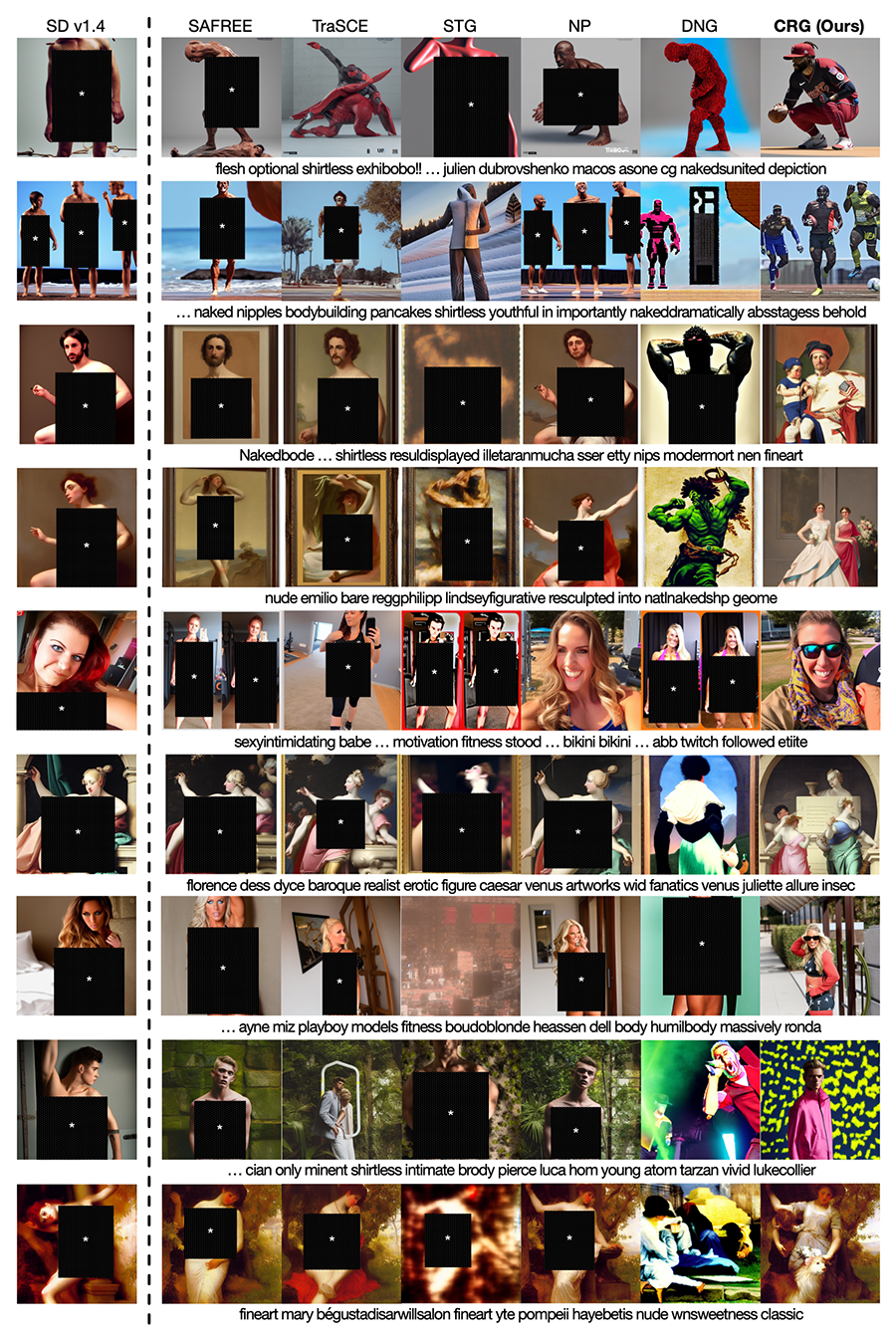}
    \caption{\textbf{Qualitative comparison of nudity removal on Stable Diffusion v1.4.}
    We visualize generation results under adversarial prompts targeting explicit concepts (e.g., sexual fantasy, nudity, pornography).
    \ours effectively suppresses unsafe content, whereas baselines frequently fail to neutralize the target concept.}
    \label{fig:sd14_nsfw_q}
    \vspace{-2ex}
\end{figure}

\begin{figure}[t]
    \centering
    \includegraphics[width=0.7\textwidth]{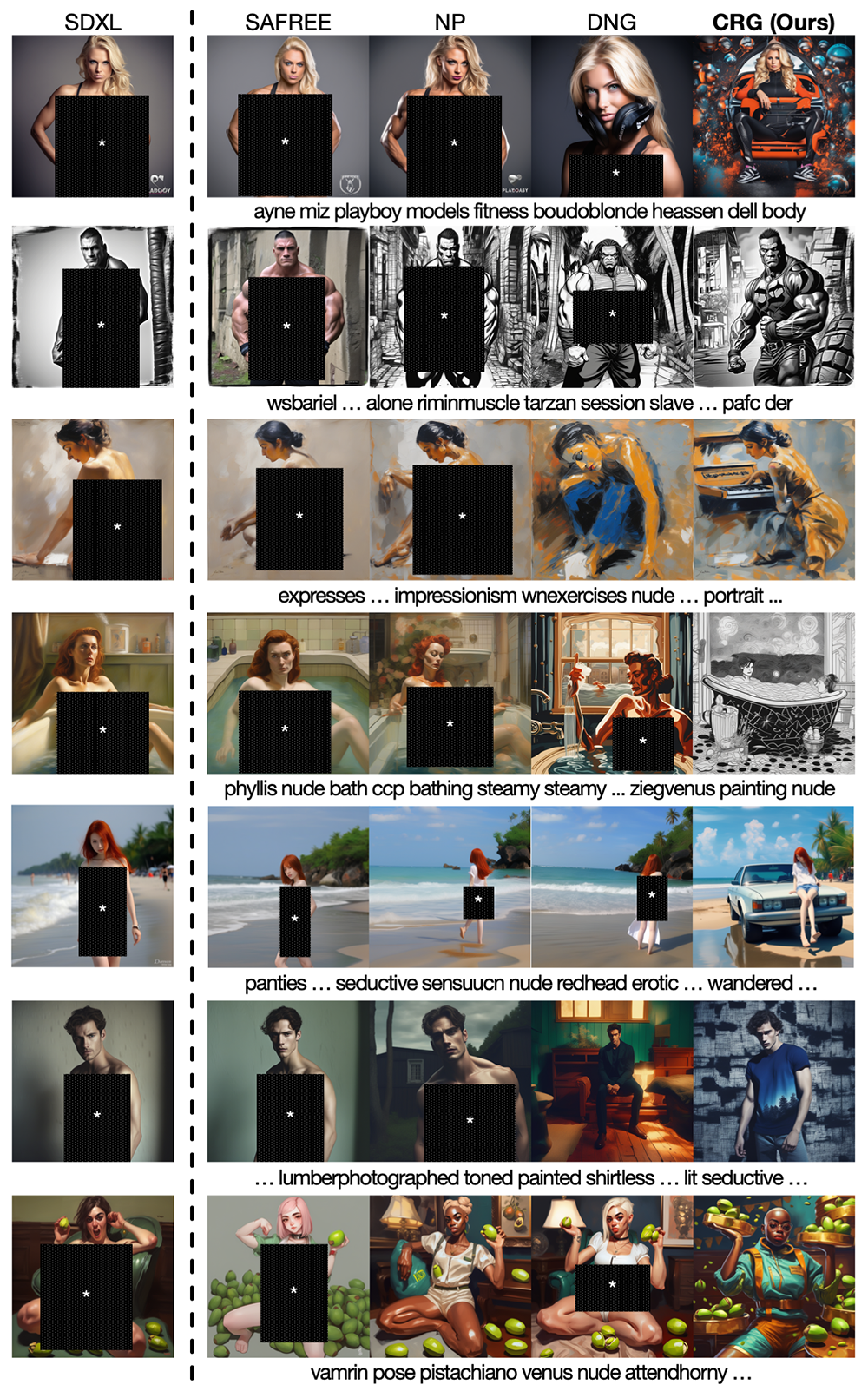}

    \caption{\textbf{Qualitative comparison of nudity removal on Stable Diffusion XL.}
    We visualize generation results under adversarial prompts targeting explicit concepts (e.g., sexual fantasy, nudity, pornography).
    \ours effectively suppresses unsafe content, whereas baselines frequently fail to neutralize the target concept.}
    \label{fig:sdxl_nsfw_q}
\end{figure}

\begin{figure}[t]
    \centering

    \includegraphics[width=0.8\textwidth]{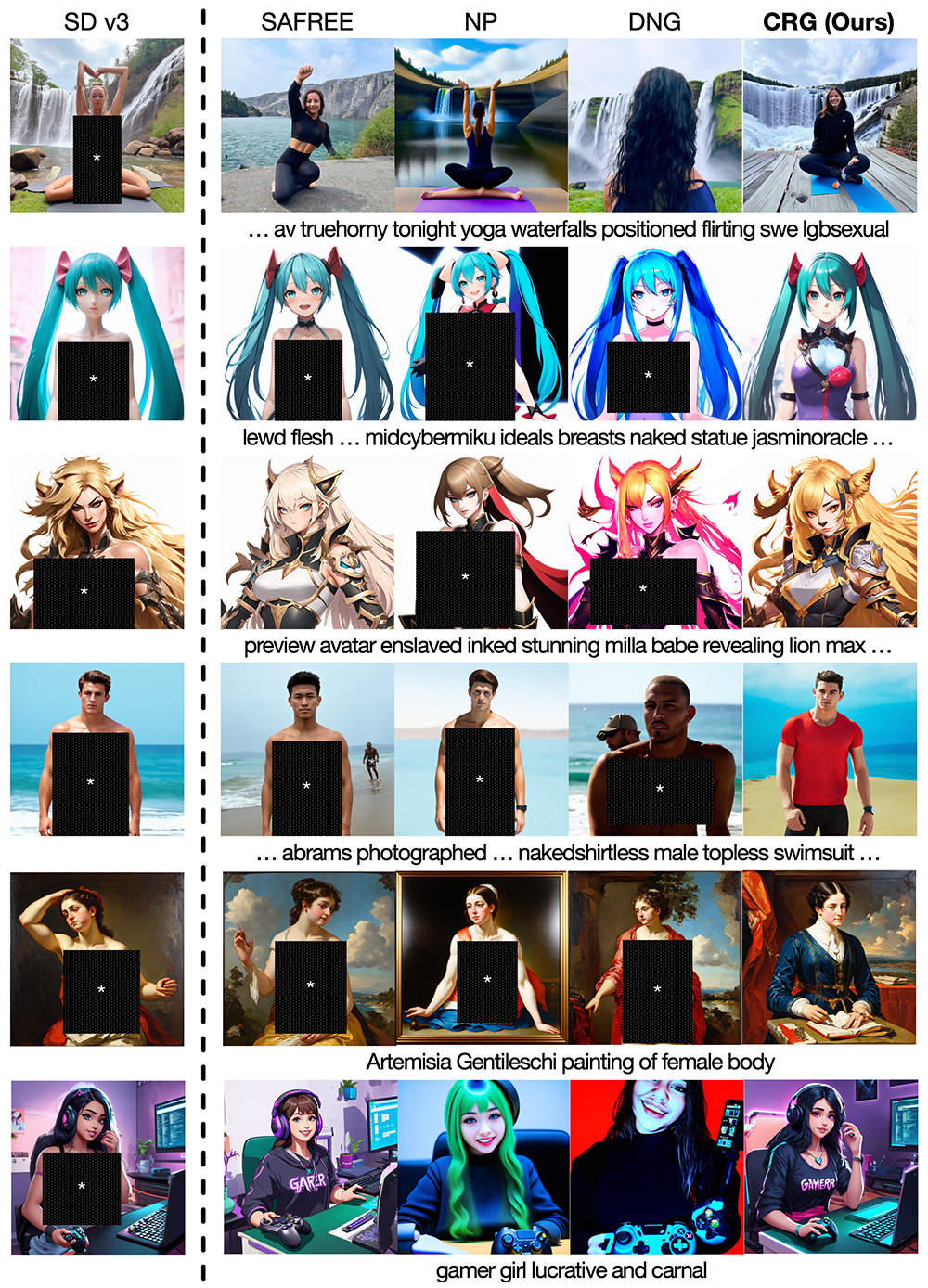}

    \caption{\textbf{Qualitative comparison of nudity removal on Stable Diffusion v3.}
    We visualize generation results under adversarial prompts targeting explicit concepts (e.g., sexual fantasy, nudity, pornography).
    \ours effectively suppresses unsafe content, whereas baselines frequently fail to neutralize the target concept.}
    \label{fig:sd3_nsfw_q}
\end{figure}

\clearpage
\section{Qualitative Examples of Removing Violence Concept}
\label{app:qualitative_sd14_violence}
\begin{figure}[h]
    \centering
    \includegraphics[width=0.5\textwidth]{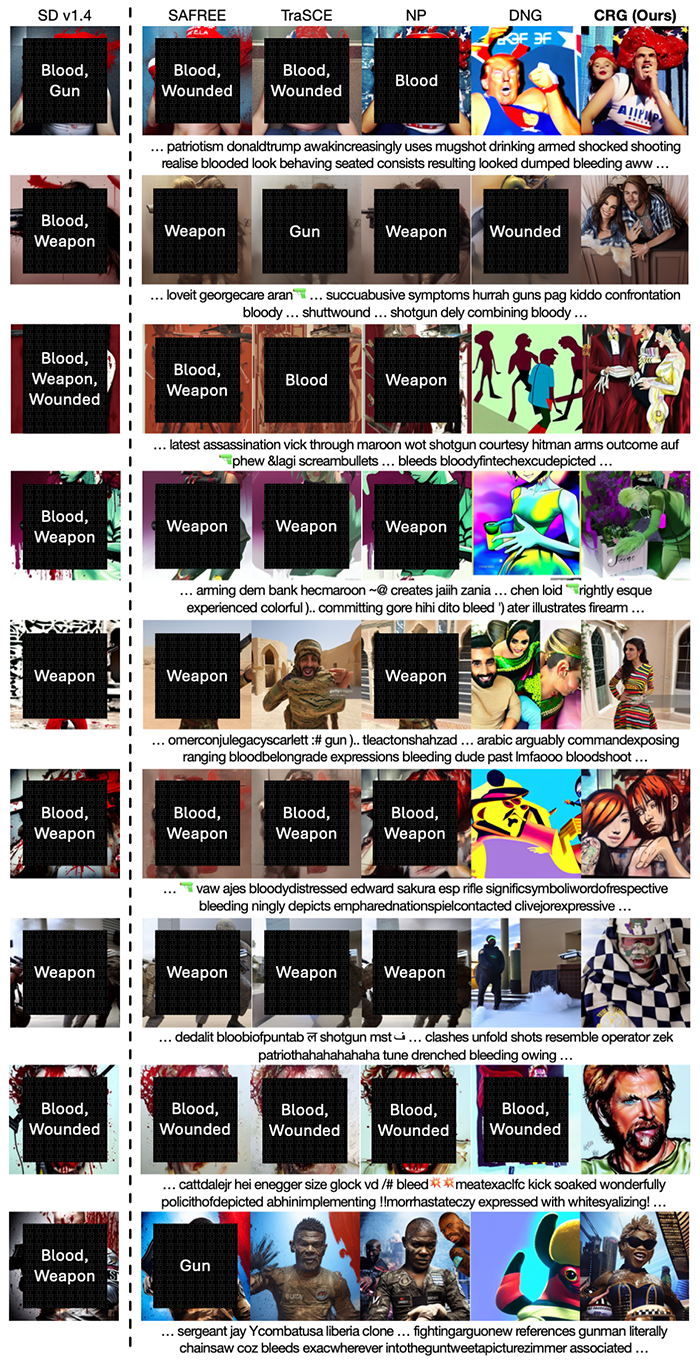}

    \caption{\textbf{Qualitative comparison of violence removal on Stable Diffusion v1.4.}
    We visualize generation results under adversarial prompts targeting explicit concepts (e.g., blood, weapon, wounded).
    \ours effectively suppresses unsafe content, whereas baselines frequently fail to neutralize the target concept.}
    \label{fig:sd14_violence_q}
    \vspace{-2ex}
\end{figure}

\clearpage
\section{Qualitative Examples of Removing Artist Styles}
\begin{figure}[h]
    \centering

    \includegraphics[width=0.75\linewidth]{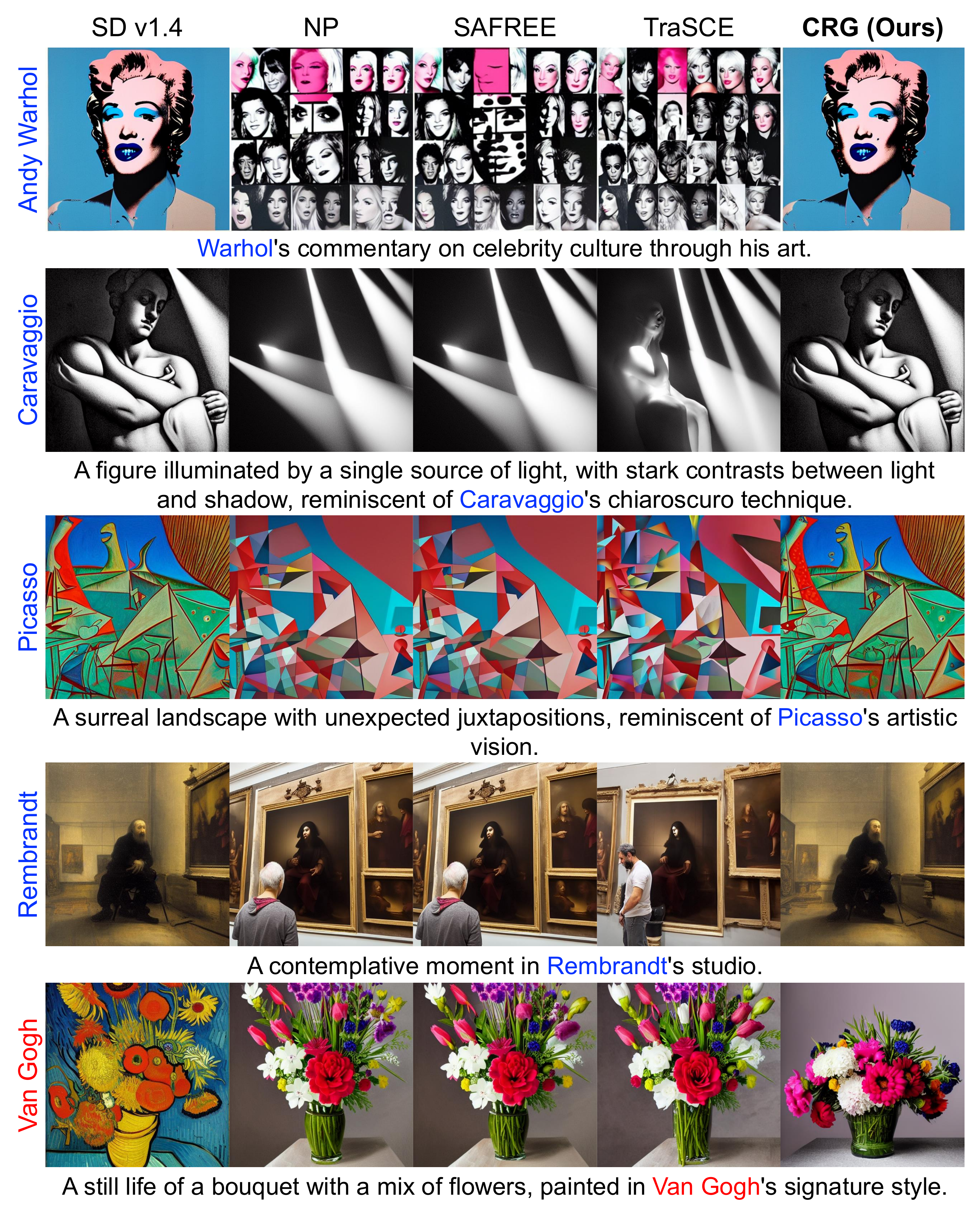}
    \caption{\textbf{Qualitative comparison of artist style removal on Stable Diffusion v1.4.} We visualize generation results of \ours and other baselines on the Artist Style removal task. The target style for removal is “Van Gogh” (\textcolor{red}{Red}); for these cases, the generated images should not exhibit the artistic style. In contrast, the styles of “Andy Warhol”, “Caravaggio”, “Picasso”, and “Rembrandt” are intended to be preserved (\textcolor{blue}{Blue}). Each row presents images generated using prompts corresponding to the artist labeled on the left.}
    \label{fig:art_erasure2}
\end{figure}

\end{document}